\documentclass[Afour,sageh,times]{sagej}

\usepackage{moreverb,url}
\usepackage[colorlinks,bookmarksopen,bookmarksnumbered,citecolor=red,urlcolor=red]{hyperref}
\usepackage{amsmath,amssymb,amsthm,bm,algorithmic,algorithm2e,xcolor,xargs}
\usepackage{graphicx}
\usepackage{multirow}
\usepackage[normalem]{ulem}
\usepackage{enumitem}
\usepackage{booktabs}
\usepackage[obeyFinal,textsize=footnotesize]{todonotes}  
\usepackage{arydshln}

\presetkeys{todonotes}{color=orange!20}{}

\newcommand{\trsp}{{\scriptscriptstyle\top}}
\newcommand{\eg}[0]{\textit{e.g.}, } 
\newcommand{\ie}[0]{\textit{i.e.}, } 

\definecolor{old_color}{RGB}{150, 150, 150}

\renewcommand{\Re}{\mathbb{R}}
\newcommand{\Ne}{\mathbb{N}}

\newcommand{\deltab}{\bm{\delta}}
\newcommand{\sigmab}{\bm{\sigma}}
\newcommand{\xb}{\bm{x}}

\newcommand{\qb}{\bm{q}}
\newcommand{\pb}{\bm{p}}
\newcommand{\kthres}{k_\mathrm{thresh}}
\newcommand{\kbinom}{k_\mathrm{binom}}
\newcommand{\krad}{k_\mathrm{rad}}
\newcommand{\kradbool}{k_\mathrm{rad}'}

\newcommand{\onebf}{\mathbf{1}}
\newcommand{\calD}{\mathcal{D}}
\newcommand{\calF}{\mathcal{F}}
\newcommand{\calH}{\mathcal{H}}
\newcommand{\calP}{\mathcal{P}}
\newcommand{\txtmin}{\mathrm{min}}

\newtheorem{proposition}{Proposition}
\newtheorem{corollary}{Corollary}
\theoremstyle{definition}
\newtheorem{definition}{Definition}
\newtheorem{assumption}{Assumption}
\theoremstyle{remark}
\newtheorem{example}{Example}
\newtheorem{remark}{Remark}

\def\volumeyear{2025}
\begin{document}

\runninghead{Bruderm\"uller et al.}

\title{CC-VPSTO: Chance-Constrained Via-Point-Based Stochastic Trajectory Optimisation for Online Robot Motion Planning under Uncertainty}

\author{Lara Bruderm\"uller\affilnum{1}, Guillaume O.~Berger\affilnum{2}, Julius Jankowski\affilnum{3}, Raunak Bhattacharyya\affilnum{1}, Sylvain Calinon\affilnum{3}, Rapha\"el M.~Jungers\affilnum{2}, and Nick Hawes\affilnum{1}} 

\affiliation{\affilnum{1}Oxford Robotics Institute, University of Oxford, UK\\
  \affilnum{2}ICTEAM, UCLouvain, Belgium\\
  \affilnum{3}Idiap Research Institute \& Ecole Polytechnique F\'ed\'erale de Lausanne (EPFL), CH
}

\corrauth{Lara Bruderm\"uller, Oxford Robotics Institute, \\23 Banbury Rd, Oxford OX2 6NN, UK
}
\email{larab@robots.ox.ac.uk}

\begin{abstract}

Reliable robot autonomy hinges on decision-making systems that account for uncertainty without imposing overly conservative restrictions on the robot's action space.
We introduce \emph{Chance-Constrained Via-Point-Based Stochastic Trajectory Optimisation (CC-VPSTO)}, a real-time capable framework for generating task-efficient robot trajectories that satisfy constraints with high probability by formulating stochastic control as a chance-constrained optimisation problem. Since such problems are generally intractable, we propose a deterministic surrogate formulation based on Monte Carlo sampling, solved efficiently with gradient-free optimisation. To address bias in naïve sampling approaches, we quantify approximation error and introduce padding strategies to improve reliability. We focus on three challenges: (i) sample-efficient constraint approximation, (ii) conditions for surrogate solution validity, and (iii) online optimisation. Integrated into a receding-horizon MPC framework, CC-VPSTO enables reactive, task-efficient control under uncertainty, balancing constraint satisfaction and performance in a principled manner.
The strengths of our approach lie in its generality, \ie no assumptions on the underlying uncertainty distribution, system dynamics, cost function, or the form of inequality constraints; and its applicability to online robot motion planning.
We demonstrate the validity and efficiency of our approach in both simulation and on a Franka Emika robot. Videos and additional material are made available \href{https://sites.google.com/oxfordrobotics.institute/cc-vpsto}{here}.
\end{abstract}


\maketitle

\section{Introduction}
Uncertainty is inherent to most real-world robotics applications, arising from noisy sensors, imprecise actuators, and incomplete or evolving knowledge of the environment. Effectively managing this uncertainty is essential for achieving reliable and efficient robot behaviour, particularly in online motion planning tasks that require fast adaptation to new information. 
In this work, we adopt a \emph{chance-constrained} perspective, where constraints such as collision avoidance (cf.~Fig.~\ref{fig:robot_exp_teaser}), force limits, or task completion cannot be guaranteed but must instead be satisfied with high probability~\citep{prekopa2013stochastic, dai2019chance}.
Unlike traditional robust control methods~\citep{kohler2023stochastic, majumdar2017funnel, badings2023robust} that optimise for the worst-case scenario under \emph{bounded uncertainty}, 
chance constraints enable a more general, \textit{probabilistic} treatment of uncertainty~\citep{margellos2014road, schildbach2014scenario}, allowing for more explicit trade-offs between constraint satisfaction and task efficiency.

Crucially, we are interested in an \textit{online} robot motion planning setting where constraint violations are undesirable but not catastrophic, and where performance (\eg motion duration) remains important. Our objective is to trade off constraint satisfaction and task performance in a principled manner that avoids unnecessary conservatism.
While Model Predictive Control (MPC) can implicitly provide some robustness via frequent replanning, it typically relies on deterministic models, leading to brittle, myopic behaviour in stochastic settings. 
Incorporating probabilistic information directly into the control loop remains a key challenge, but is essential for enabling more flexible and robust decision-making in uncertain environments.

Chance-constrained formulations, which require that constraints must be satisfied with high probability (\eg at least 95\%), offer a natural solution but are in general intractable~\citep{blackmore2010probabilistic}. 
One common strategy is to approximate the chance constraint and reformulate the problem as a deterministic surrogate that can be addressed using standard optimisation techniques.
However, identifying a suitable approximation is often non-trivial. Common approaches either introduce significant conservatism at the cost of task efficiency~\citep{lew2023risk, calafiore2006scenario}, or, like naïve sample average approximation (SAA) methods~\citep{shapiro2021lectures, shapiro2003monte, pagnoncelli2009sample}, are overly optimistic under limited samples, biasing solutions toward regions that appear feasible in the surrogate problem but violate the true constraint~\citep{homem2014monte}, a phenomenon we also observed in our experiments (cf., \eg Fig.~\ref{fig:constraint-type-vs-samples}).

Towards this end, we propose \textbf{CC-VPSTO} (Chance-Constrained Via-Point-Based Stochastic Trajectory Optimization), a novel Monte Carlo-based framework for \textit{online} robot motion planning \textit{under uncertainty}. CC-VPSTO systematically balances the inherent trade-offs between probabilistic constraint satisfaction, solution quality, and computational efficiency by focusing on three key challenges: \textit{a)} given a number of samples and a \emph{confidence level}, selecting a statistical \emph{padding}, \ie a margin on the empirical constraint, that ensures the sample-based solution satisfies the true chance constraint with high confidence; \textit{b)} keeping the padding small to avoid excessive conservatism; and \textit{c)} solving the problem efficiently in real time with few samples. Let us emphasize that \emph{a)}, \emph{b)}, and \emph{c)} are competing objectives since few samples usually result in overly conservative padding (such as in scenario optimisation approaches; \eg~\citealp{calafiore2006scenario}) or overly optimistic solutions (such as in naïve sample average approximation approaches; \eg~\citealp{shapiro2021lectures}). 
In summary, the main contributions of this work are:

\begin{figure}[t!]
    \centering
    \includegraphics[width=0.9\columnwidth]{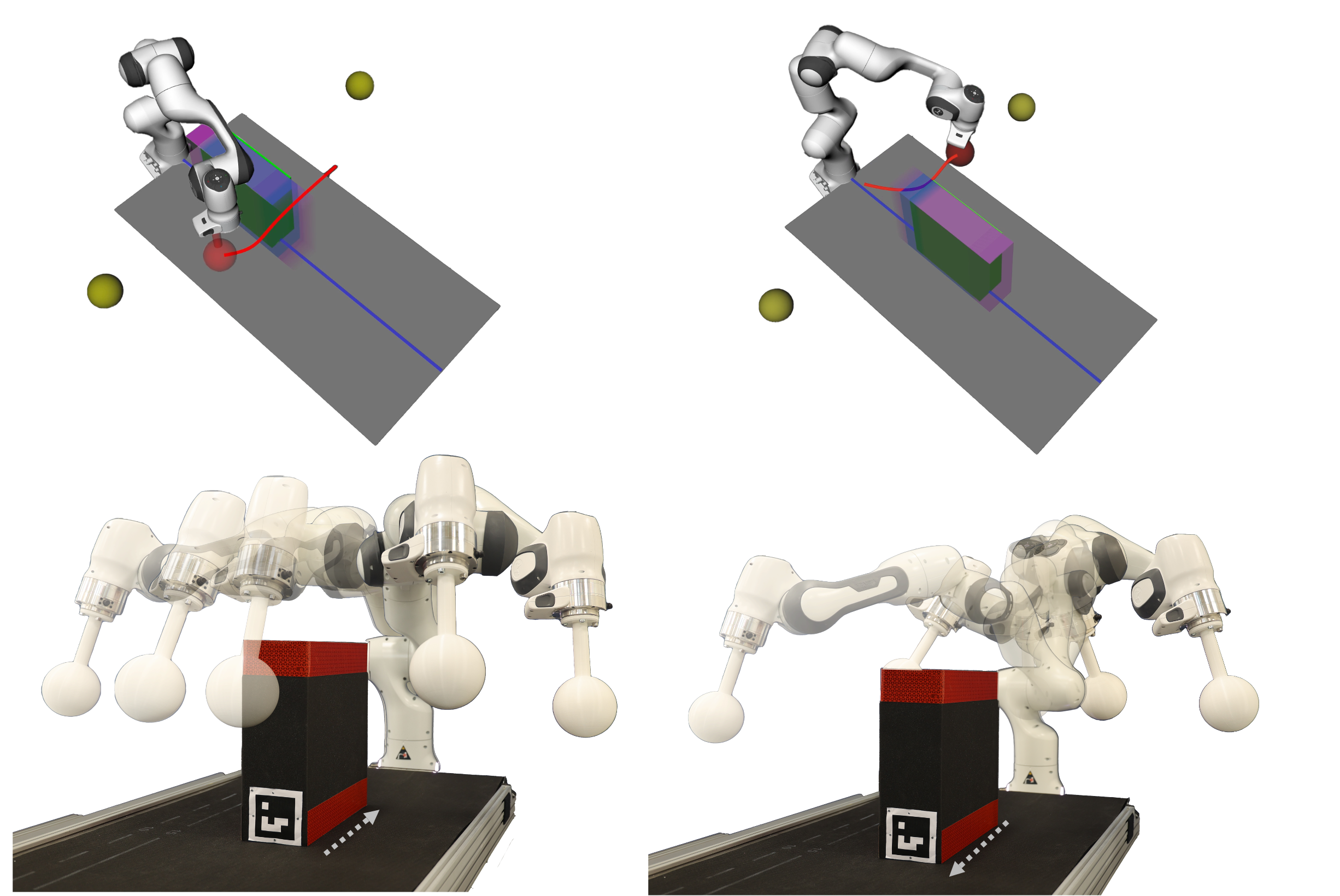}
    \caption{\emph{\textbf{Robot experiment.}}
    The robot must move its ball-shaped end effector from a start to a goal point (yellow spheres) across a conveyor belt, while avoiding a box obstacle moving stochastically on the belt. Depending on the anticipated box motion, the robot can pass in front (left) or behind (right). This task requires online, reactive motion generation that balances constraint satisfaction with task efficiency, aiming to reach the goal as quickly as possible under uncertainty.}
    \label{fig:robot_exp_teaser}
\end{figure}

\begin{enumerate}[leftmargin=*]
    \item A new surrogate formulation for chance-constrained trajectory optimisation, designed specifically for online motion planning. It maintains low added conservatism even with small sample sizes, while accounting for approximation errors to avoid overly optimistic solutions.
    
    \item A theoretical analysis of the surrogate’s correctness. Under the assumption of independence between the solution and the samples, we show that the solution of the surrogate problem satisfies the original (intractable) chance constraint with high confidence. We further provide insights into why the approach remains effective in receding-horizon settings, such as MPC, where the independence assumption may not strictly hold.
    
    \item The empirical evaluation of our approach across a range of challenging tasks. Even when used heuristically (i.e., without guaranteed independence), the surrogate performs reliably in both simulation and real-world robot experiments.

    \item The integration of the surrogate formulation into \textit{VP-STO}, an MPC framework for \textit{online} reactive robot control~\citep{jankowski2022vp}, extending it to a stochastic setting. This enables receding-horizon control that effectively balances constraint satisfaction and task performance under uncertainty.

\end{enumerate}

The key advantages of our approach are: \textit{i)} flexibility to handle arbitrary uncertainty distributions, \textit{ii)} compatibility with real-time MPC via parallelisable sampling and optimisation, and \textit{iii)} support for general inequality constraints, such as collision avoidance, force limits, or performance objectives.

\section{Related Work}\label{sec:related_work}
Risk-averse planning and control methods in robotics aim to enforce constraints in the presence of uncertainty. These methods typically enforce these constraints by formulating them as chance constraints (CCs) or by using risk measures, such as Conditional Value-at-Risk (CVaR). 
When employed in a \emph{online} receding horizon control scheme, these methods fall into the category of Stochastic Model Predictive Control (SMPC) \citep{heirung2018stochastic, mesbah2016stochastic}. SMPC addresses optimal control problems for dynamical systems under stochastic uncertainty, while enforcing constraints that must be satisfied with high probability (\ie chance constraints).
In the context of chance-constrained control, the literature distinguishes  between two types of formulations for limiting constraint violations. These can be defined either \emph{point-wise}, \ie imposing a separate probability bound at each time step, or \emph{jointly}, where the constraint must hold over the entire (finite) time horizon with high probability.
We focus on joint chance constraints in this work, as they are preferable in robotics where it is important to account for the cumulative effect of uncertainty over time.
As discussed by \citet{janson2017monte}, many approaches bound the joint probability using either an \emph{additive} approach, summing over all point-wise probabilities via Boole's inequality (\eg \citet{ono2008iterative, priore2023chance, castillo2020real}), or a \emph{multiplicative} approach, which involves explicitly constraining the product of the complements of point-wise probabilities (\eg \citet{sun2016safe, van2011lqg}). Yet, both of these strategies do not account for the time correlations of uncertainty, and thus may lead to over- or underestimation of the joint probability of constraint violation. We thus adopt the time-wise supremum approach from~\citet{lew2023risk}, which evaluates only the maximum constraint violation over the entire time horizon for a given trajectory, thereby capturing time correlations effectively.

The main challenge in SMPC is evaluating the probability of constraint violation over the planning horizon. This requires computing an expectation integral over time and space, which is typically intractable for general uncertainty distributions and constraint structures~\citep{pena2020solving}. As a result, SMPC must address two key questions: \textit{i)} how to approximate or bound the probability of constraint violation in a tractable way, and \textit{ii)} how to solve the resulting optimisation problem with minimal computational overhead for online control.
Previous approaches to these questions proposed semidefinite programming formulations \citep{jasour2015semidefinite} or constraint tightening \citep{alcan2022differential, ono2008iterative, parsi2022computationally}.
While effective in providing probabilistic guarantees on the satisfaction of chance constraints, they are typically tailored to very specific types of constraints, uncertainty distributions and/or system dynamics, thereby limiting their applicability to real robotics problems. As noted in \citet{lew2023risk}, there is still a lack of formulations and solution algorithms that are capable of capturing different sources of uncertainty as well as different types of constraints in a unified framework. 

In contrast, sample-based methods offer a more general approach for approximating chance constraints, as they do not require any assumptions on the underlying probability distributions, as long as the number of samples is sufficiently large. 
In the sample-based setting, we can distinguish between \emph{scenario optimisation} \citep{schildbach2014scenario, de2023scenario} and \emph{Monte Carlo} methods \citep{blackmore2010probabilistic, schmerling2016evaluating, blackmore2006particle}.
Both approaches use samples (aka.~scenarios) to capture the underlying uncertainty.
Scenario optimisation synthesises controls satisfying the constraint for each of the samples and relies on a well-established theory to identify the right sample size for a given confidence level~\citep{calafiore2006scenario}.
However, these theoretical bounds are mostly limited to convex or quasi-convex problems \citep{calafiore2010random, berger2021chance} and solutions are typically overly conservative, \ie they require much larger sample sizes than identified by empirical tests \citep{schildbach2014scenario}.
Monte Carlo methods typically approximate the probability of constraint violation from the samples, rooted in the \emph{sample average approximation (SAA)} approach \citep{shapiro2021lectures, shapiro2003monte, pagnoncelli2009sample}.  
They are generally less conservative and can be used with arbitrary constraints and uncertainty models.
However, without further adjustments, they do not provide finite-sample guarantees, but only asymptotic guarantees \citep{blackmore2006particle}, implying the requirement of large sample sets, and higher computational resources.
The need for large sample sets is reinforced when the desired probability of constraint violation is low, as is commonly targeted in robotics applications.
A remedy to this can be importance sampling \citep{schmerling2016evaluating}, or data reduction methods based on parameter estimation of sample statistics, \eg through computing moments of the probability distribution of the uncertainty \citep{wang2020moment, priore2023chance, blackmore2006probabilistic, yan2018stochastic}.
Yet, the propagation of moments can be complex, and requires restrictive assumptions, such as Gaussianity.
Alternatively, for collision avoidance, \citet{trevisan2025dynamic} propose a naïve Monte Carlo approach that approximates the chance constraint using a fixed number of samples, increasing sample density by constraining the collision region across time steps. However, this method is tailored specifically to collision avoidance and does not account for the approximation error introduced by the finite sample size.

Other approaches have used SAA to approximate constraints on risk metrics like conditional Value-at-Risk (CVaR) instead~\citep{lew2023risk,yin2023risk, nemirovski2007convex}. CVaR constraints are more conservative than chance constraints, as they account for tail events, but the resulting reformulation is smooth and convex, which enables the use of off-the-shelf optimisation tools, such as sequential convex programming (SCP). For instance, \citet{lew2023risk} provide a general framework for risk-averse SMPC based on the combination of the SAA of CVaR constraints and concentration inequalities to bound the approximation error. However, their approach relies on strong continuity assumptions on the objective function and constraints, which may not hold in practice. \citet{yin2023risk} use the SAA of CVaR constraints within a Model Predictive Path Integral (MPPI) controller but do not account for errors in the approximation of the CVaR constraint and limit the source of uncertainty to process noise. 
Finally, the work of \citet{pena2020solving} reformulates the chance constraint as a quantile function and uses SAA to approximate it, which results in a formulation that is amenable to gradient-based optimisation methods. However, similar to \citet{lew2023risk}, their approach relies on continuity and differentiability assumptions of the constraint function. 

Our main observation across chance constrained optimisation approaches is that they are typically tailored to specific constraint models (\eg collision avoidance constraints, or reaching polytopic target sets) with smoothness and continuity assumptions, or specific types of uncertainty (\eg Gaussian or bounded uncertainty). These restrictive assumptions limit the applicability of these approaches in real-world robotics problems.
In contrast, we provide a general framework for chance-constrained finite-horizon optimal control problems with generic chance constraints and generic uncertainty distributions. Building upon the SAA approach, we propose a sample-based approximation that actively accounts for the approximation error caused by the number of samples used in the approximation by adjusting the threshold for constraint violation based on a fixed confidence level. This allows us to use a small number of samples in order to efficiently solve the resulting deterministic problem with the stochastic trajectory optimisation framework VP-STO \citep{jankowski2022vp}.

\section{Problem Formulation}

%

\subsection{Preliminaries}

Let $\Delta$ be a random variable that models the uncertainty of the system.
This can include stochasticity in the dynamics, the environment, and the sensor measurements.
A realization $\deltab$ of $\Delta$, denoted by $\deltab\sim\Delta$, will be referred to as a \emph{sample}, or a \emph{scenario} of the uncertainty.

\subsection{Chance-Constrained Optimisation}

In the following, we introduce the general chance-constrained optimisation problem. 
The goal is to find a solution $\xb$ that minimizes a cost $J(\xb){} \in \Re$ while satisfying a set of constraints.
In our work, we consider inequality constraints, \ie constraints that can be formulated as a function $g$ being negative at $\xb$, \ie $g(\xb)\leq0$.
For instance, $g$ can encode a deterministic collision-avoidance constraint on the robot's distance to obstacles.

Chance-constrained optimisation generalises the above by allowing constraints that depend on a random variable.
More precisely, the constraints have the form $g(\xb,\deltab)\leq0$, where $g$ depends on $\xb$ and the realization $\deltab$ of the uncertainty variable $\Delta$.
For instance, $g(\xb,\deltab)\leq0$ can encode a collision-avoidance requirement  of a stochastic system in state $\bm{x}$ given a particular uncertainty realisation $\deltab$.
Requiring that $g(\xb,\deltab)\leq0$ holds for \emph{all} realizations of $\deltab$ is often overly conservative, or even infeasible.
This is especially true if the distribution of $\Delta$ has unbounded support (such as Gaussian noise).
Therefore, chance-constrained optimisation relaxes the constraint into a soft constraint, allowing violation of the constraint with a bounded probability $\eta$. It thus requires that the probability of a realisation $\deltab\sim\Delta$ to satisfy $g(\xb,\deltab)>0$ is smaller than $\eta$.
A general chance-constrained optimisation problem can be formulated as follows:
\begin{equation}\label{eq:cc_opt_basic}
\begin{split}
    \min_{\xb\in X} \quad & J(\xb) \\
    \text{s.t.} \quad & P_{\deltab\sim\Delta}[g(\xb, \deltab) > 0] \leq \eta,
\end{split}
\end{equation}
where $\xb$ is the decision variable, constrained in some domain $X$ (e.g., $X\subseteq\Re^n$), $J:X\rightarrow\Re$ is the objective function, and $\eta\in[0, 1]$ is a user-provided threshold for the probability of violating $g$.
We assume that the probability distribution of $\Delta$ is known or that we have a generative model for $\Delta$ from which we draw samples, \ie we can draw an arbitrary number of \emph{independent} samples $\deltab_1\sim\Delta,\ldots,\deltab_N\sim\Delta$.
As mentioned above, the chance constraint is satisfied at $\xb$ if the probability of violating the constraint $g$ at $\xb$ is at most $\eta$.
Computing this probability for a given $\xb$ is often challenging.
For this reason, chance-constrained optimisation problems are often very hard, if not impossible, to solve exactly.
To this end, we contribute a tractable approximation of such problems, along with an analysis of the soundness of the approximation.

\begin{remark}
Note that Eq.~\eqref{eq:cc_opt_basic} can be generalized to multiple chance constraints $P_{\deltab\sim\Delta}[g_i(\xb,\deltab)>0]\leq\eta_i$, for $i=1,\ldots,L$, with different violation thresholds $\eta_i$.
However, in this work, we will focus on a single joint constraint ($L=1$) for simplicity.
\end{remark}

Note, that we do not make any additional assumptions on the uncertainty distribution, \ie it can be of any type and is not restricted to additive noise formulations. 
However, note that state-dependent uncertainties are outside the scope of this work. The following example illustrates possible sources of uncertainty in a simplified robot motion planning problem.

\begin{example}[\textit{System with uncertain initial condition, actuation noise and uncertain obstacle dynamics}]\label{exa:simple-system}
Consider a simple kinodynamic system in discrete time: $s(t+1)=s(t)(u(t)+w(t))$,where \( s(t) \in \mathbb{R} \) is the system state, \( u(t) \in \mathbb{R} \) the control input, \( w(t) \in \mathbb{R} \) the actuation noise, and \( z(t) \in \mathbb{R} \) the position of a randomly moving obstacle that occupies the region \( [z(t), \infty) \).  
The random variables \( s(0) \), \( w(t) \), and \( z(t) \) follow known distributions; for instance, \( s(0) \sim \mathcal{N}(0, 1) \), \( w(t) \sim \mathcal{U}(-1, 1) \), and \( z(t) \sim \mathrm{Exp}(\lambda) \) for all \( t \).
The objective is to minimize the sum of squared inputs over two time steps, while avoiding the obstacles with high probability over two time steps.
Following the formulation of \eqref{eq:cc_opt_basic}, we have $\xb=(u(0),u(1))$, $\deltab=(s(0),w(0),w(1),z(0),z(1),z(2))$, $J(\xb)=u(0)^2+u(1)^2$, and $g(\xb,\deltab)=\max_{t=0,1,2}\,s(t)-z(t)$, where $s(t)$ follows the dynamics introduced above.
\end{example}%

\subsection{Constraint Satisfaction as a Binary Random Variable}
In this subsection, we reformulate the chance constraint in Eq.~\eqref{eq:cc_opt_basic} as a constraint on a binary random variable obtained from $\Delta$.
The motivation for doing this is that we will use this formulation to define a \emph{sample average approximation} of the chance constraint in the next section.
Concretely, given $\xb$, we introduce a binary random variable $G_{\xb}=\boldsymbol{1}_{g(\xb,\Delta)>0}$, wherein $\boldsymbol{1}_{(\cdot)}$ denotes the indicator function, \ie for any $\deltab$ sampled from $\Delta$, $G_{\xb}=1$ if $g(\xb,\deltab)>0$; otherwise, $G_{\xb}=0$.


$G_{\xb}$ is a random variable since it depends on the random uncertainty variable $\Delta$.
Thus, we are interested in the probability distribution of the value of $G_{\xb}$.
By definition, this can be obtained from the probability distribution of $\Delta$: namely, $P[G_{\xb}=1]=P_{\deltab\sim \Delta}[g(\xb,\deltab)>0]$.
Hence, the chance constraint in Eq.~\eqref{eq:cc_opt_basic} can be rewritten as 
\begin{align}\label{eq:cc_binom_version}
P[G_{\xb}=1] \leq \eta.
\end{align}

\begin{remark}\label{rem:integral}
If we know the probability density function $p_\Delta$ of $\Delta$, then $P[G_{\xb}=1]$ can be obtained by computing the integral
\begin{equation}\label{eq:total_prob}
P[G_{\xb}=1]=\int_\calD \boldsymbol{1}_{g(\xb,\deltab)>0}\ p_\Delta(\deltab) \; \mathrm{d}\deltab,
\end{equation}
where the integration domain $\calD$ consists of all realizations $\deltab$ of $\Delta$.
However, computing this integral is generally intractable in practice, especially when the dimension of $\Delta$ is large.
\end{remark}%

\section{Sample Average Approximation}\label{sec:monte_carlo}

Because of the challenges in computing the probability in Eq.~\eqref{eq:cc_binom_version} exactly (see Remark~\ref{rem:integral}), a tractable approximation is required. 
A popular approach is the \textit{particle-based} approximation proposed by \citet{blackmore2010probabilistic}.
The core concept is to draw a finite set of i.i.d.\ uncertainty samples, or samples, and approximate $P[G_{\xb}=1]$ as an average over the samples.
This approach is justified by the law of large numbers: as the number of samples tends to infinity, the average converges to $P[G_{\xb}=1]$.
In the remainder of this work, we refer to this as a \textit{sample average approximation}~\citep{pagnoncelli2009sample}, also known as a Monte Carlo approximation, but we adopt the SAA terminology to highlight its use within an optimisation context.
 
Formally, consider a set $D=\{\deltab_i\}_{i=1}^N$ of $N$ i.i.d.~samples drawn from $\Delta$.
Based on $D$, a sample average approximation of $P[G_{\xb}=1]$ can be computed as follows:

\begin{equation}\label{eq:approx}
P[G_{\xb}=1] \approx \frac{1}{N} \underbrace{\sum_{i=1}^N \boldsymbol{1}_{g(\xb,\deltab_i)>0}}_{s_N(\xb;D)}.
\end{equation}

This is equivalent to counting the number $s_N(\xb;D)$ of samples $\deltab_i$ in $\{\deltab_i\}_{i=1}^N$ that violate the constraint $g$ at $\xb$ and dividing it by the total number of samples $N$.
Note that given $\xb$ and $\deltab$, determining whether $g(\xb,\deltab)\leq0$ and computing $s_N(\xb;D)$ for a given solution $\xb$ is generally much cheaper than computing the integral in Eq.~\eqref{eq:total_prob}.

We can now use the approximation in Eq.~\eqref{eq:approx} to construct a surrogate constraint of the intractable chance constraint in the optimisation problem Eq.~\eqref{eq:cc_opt_basic}.
A naïve approach is to simply replace $P[G_{\xb}=1]$ in Eq.~\eqref{eq:cc_binom_version} by its approximation, \ie require that $\frac1Ns_N(\xb;D)\leq\eta$, or equivalently that $s_N(\xb;D)\leq\eta N$. 
By the law of large numbers, when the number of samples approaches infinity, the feasible set of this surrogate constraint asymptotically converges to that of the original chance constraint (cf. Eq.~\eqref{eq:cc_binom_version}).
However, when using a finite number of samples, satisfaction of the surrogate constraint does not guarantee that the original chance constraint is satisfied.
The reason is that we need to account for the approximation error in Eq.~\eqref{eq:approx}.
This can be achieved through strengthening the surrogate constraint: by requiring that
\begin{equation}\label{eq:surrogate-constraint}
s_N(\xb;D)\leq\kthres
\end{equation}
for some $\kthres<\eta N$.
The precise value of $\kthres$ depends on two parameters: \textit{i)} the number of samples $N$, and \textit{ii)} the level of \emph{confidence} that we want on the \emph{soundness} of the surrogate constraint in Eq.~\eqref{eq:surrogate-constraint}.
Determining suitable values for $\kthres$ is the main contribution of this paper.

Leveraging Eq.~\eqref{eq:surrogate-constraint}, we formulate the following surrogate optimisation problem to the original chance-constrained optimisation problem Eq.~\eqref{eq:cc_opt_basic} as follows: 
\begin{equation}\label{eq:cc_opt_surrogate}
\begin{split}
    \min_{\xb\in X} \quad & J(\xb) \\
    \text{s.t.} \quad & s_N(\xb;D) \leq \kthres.
\end{split}
\end{equation}
Our goal is to determine values of $\kthres$ (as a function of $N$ the number of samples) such that feasible solutions of Eq.~\eqref{eq:cc_opt_surrogate} are feasible for the original problem Eq.~\eqref{eq:cc_opt_basic} with user-given \emph{confidence}.

The term \emph{confidence} above refers to the probability that we sample $N$ samples $D=\{\deltab_i\}_{i=1}^N$ for which satisfying the surrogate constraint Eq.~\eqref{eq:surrogate-constraint} implies satisfaction of the original chance constraint in Eq.~\eqref{eq:cc_binom_version}.
Although the highest confidence of $1$ ($100\,\%$) would be desirable, this in general only achievable in the limit, \ie when $N\to\infty$.
Indeed, when $N$ is finite, there is in general a non-zero probability of sampling a set of samples, such that the true chance constraint Eq.~\eqref{eq:cc_binom_version} may be violated even though the surrogate constraint Eq.~\eqref{eq:surrogate-constraint} is satisfied.
However, we can leverage the confidence to establish a connection between the number of samples $N$ and the threshold $\kthres$ in the surrogate constraint Eq.~\eqref{eq:surrogate-constraint} to account for the approximation error arising from the finite number of samples.

\subsection{Sample Average Approximation as a Bernoulli Process}
The approximation Eq.~\eqref{eq:approx} of $P[G_{\xb}=1]$ can be interpreted as a Bernoulli process, \ie the act of drawing $N$ independent samples from a given binary random variable $G$.
This connection allows us to derive suitable values for $\kthres$ as a function of $N$ and the confidence $1-\beta$, which we will discuss in the subsequent Secs.~\ref{sec:confidence-bounded-surrogate} and~\ref{sec:limitation-bernoulli}.

\paragraph*{Bernoulli process:}
is a sequence of $N$ i.i.d.~binary random variables $G_1,\ldots,G_N$, where each variable follows the same Bernoulli distribution with success probability $p$, denoted as $G_i \sim \mathrm{Bern}(p)$\footnote{Note that in the more general definition, the sequence of a Bernoulli process can also be infinite.}. Hence, every variable in the sequence is associated with a Bernoulli trial that has a binary outcome governed by the Bernoulli distribution $\mathrm{Bern}(p)$. The resulting sum of the outcomes of the Bernoulli trials, \ie $S_N=\sum_{i=1}^N G_i$, is a random variable that follows a \emph{binomial distribution} \citep{taboga2017lectures}, \ie for all $k=0,\ldots,N$,

\begin{equation}\label{eq:binomial}
P[S_N=k]=\binom{N}{k}p^k(1-p)^{N-k},
\end{equation}
where $p=P[G=1]$.

In the sample average approximation Eq.~\eqref{eq:approx}, the binary random variable $G$ is $G_{\xb}$ and the corresponding Bernoulli trials are given by $\boldsymbol{1}_{g(\xb,\deltab_i)>0}$, $\deltab_i\sim\Delta$, for each $i=1,\ldots,N$.
Since with a fixed $\xb$ and $\{\deltab_i\}_{i=1}^N$ being i.i.d., the trials are independent, it holds that for all $k=0,\ldots,N$,
\[
P_{\deltab_1\sim \Delta,\ldots,\deltab_N\sim \Delta}[s_N(\xb;D)=k]=\binom{N}{k} p^\ell (1-p)^{N-k},
\]
where $D=\{\deltab_i\}_{i=1}^N$ and $p=P[G_{\xb}=1]$.

The above yields a closed-form expression of \emph{confidence} through the \textit{cumulative distribution function (CDF)} $C(k;N,p)$ of the binomial distribution with parameters $N$ and $p$, defined for all $k=0,\ldots,N$ by
\begin{equation}\label{eq:cdf}
    C(k;N,p) = \sum_{\ell=0}^{k} \binom{N}{\ell} p^\ell (1-p)^{N-\ell}.    
\end{equation}
\subsection{Confidence-Bounded Surrogate Constraint}\label{sec:confidence-bounded-surrogate}

In the following, we leverage the Bernoulli formulation of the sample average approximation in Eq.~\eqref{eq:surrogate-constraint} and the closed-form expression of the CDF in Eq.~\eqref{eq:cdf} to determine a threshold $\kthres=\kbinom(\beta,N,\eta)$\footnote{When clear from the context, we will drop the arguments $\beta$, $N$ and $\eta$ in $\kbinom(\beta,N,\eta)$.}. We set this threshold such that the true chance constraint Eq.~\eqref{eq:cc_binom_version} is satisfied with a user-defined \emph{confidence} $1-\beta\in(0,1)$ (where typically, $\beta\ll1$). 
This confidence level applies to any solution $\xb$ that adheres to the surrogate constraint 
\begin{equation}\label{eq:surrogate}
s_N(\xb;D)\leq\kbinom(\beta,N,\eta).
\end{equation}
on the sampled set of uncertainty samples $D=\{\deltab_i\}_{i=1}^N$.
We refer to Eq.~\eqref{eq:surrogate} as the \emph{confidence-bounded surrogate constraint} to the original chance constraint in Eq.~\eqref{eq:cc_binom_version}.
In addition, for simplicity of notation, we also define $\eta_{\mathrm{binom}}=\frac1N\kbinom$.

\begin{proposition}\label{prop:cc_binom}
Let $\beta\in(0,1)$, $N\in\Ne_{>0}$, $\eta\in(0,1)$ and let
\begin{equation}\label{eq:confidence}
\kbinom(\beta,N,\eta) = \max\big\{k \in \mathbb{N} \mid C(k;N,\eta)\leq\beta \big\}
\end{equation}
Let $\xb_{\mathrm{reject}}$ be a solution that violates the chance constraint in Eq.~\eqref{eq:surrogate}, \ie such that $P[G_{\xb_{\mathrm{reject}}}=1]>\eta$.
It holds that
\begin{equation}\label{eq:reject}
P_{\deltab_1\sim\Delta,\ldots,\deltab_N\sim\Delta}[s_N(\xb_{\mathrm{reject}};D)>\kthres] \geq 1-\beta,
\end{equation}
where $D=\{\deltab_i\}_{i=1}^N$ and $\kthres=\kbinom(\beta,N,\eta)$.
\end{proposition}
The inequality in Eq.~\eqref{eq:reject} is a lower bound on the probability of \emph{correctly rejecting} a candidate solution $\xb$ by means of the surrogate constraint in Eq.~\eqref{eq:surrogate}.

\begin{proof}
Given $\xb_{\mathrm{reject}}$ as in the proposition, we look at the probability that $s_N(\xb_{\mathrm{reject}};D)\leq \kthres$, \ie the probability that we do not reject $\xb_{\mathrm{reject}}$ when using the surrogate constraint in Eq.~\eqref{eq:surrogate}.
Since $G_{\xb_{\mathrm{reject}}}$ is a binary random variable with probability $p=P[G_{\xb_{\mathrm{reject}}}=1]$ and $D=\{\deltab_i\}_{i=1}^N$ are independent, the sum $s_N(\xb_{\mathrm{reject}};D)$ follows a binomial distribution with parameters $N$ and $p$, for which the CDF is given by Eq.~\eqref{eq:cdf}.
This implies that the probability that $s_N(\xb_{\mathrm{reject}};D) \leq \kbinom$ is equal to $C(\kbinom;N,p)$.
We build on the fact that the binomial distribution is monotonic with respect to $p$ \citep{taboga2017lectures}, \ie $p_1<p_2$ implies $C(k;N,p_1)>C(k;N,p_2)$.
Hence, it holds that $P_{\deltab_1\sim\Delta,\ldots,\deltab_N\sim\Delta}[s_N(\xb_{\mathrm{reject}};D) \leq \kbinom]$ is smaller than or equal to $C(\kbinom;N,\eta)$ since $p>\eta$.
Now, by definition of $\kbinom$, it holds that $C(\kbinom;N,\eta)\leq\beta$, so that the probability of not rejecting $\xb_{\mathrm{reject}}$ is at most $\beta$.
\end{proof}

\begin{figure}[t]
    \centering
    \includegraphics[width=0.9\linewidth]{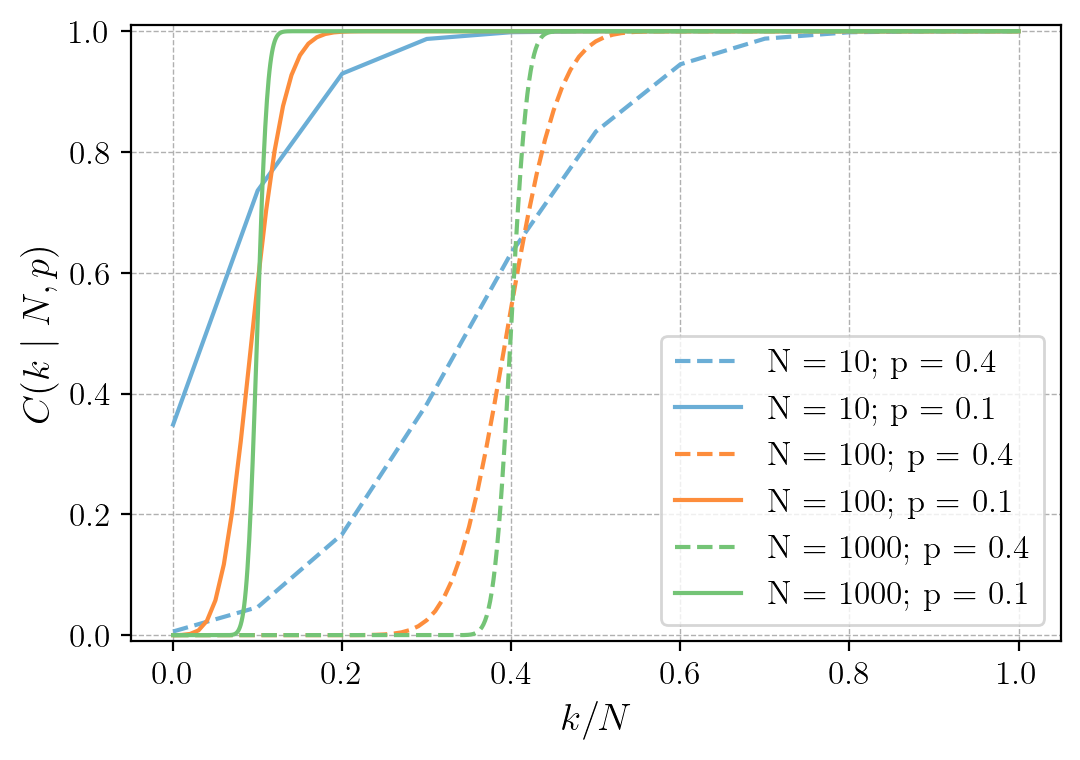}
    \vspace{-0.4cm}
    \caption{CDF of the binomial distribution for different values of $N$ and $p$, given $k/N$ on the x-axis. The CDF values map to our confidence in observing $k/N$ constraint violations under the assumption that the true probability of constraint violation is $p$.}
    \label{fig:cdf_against_eta}
\end{figure}

In Fig.~\ref{fig:cdf_against_eta}, we show the CDF of the binomial distribution for different values of $N$ and $p$.
From the plots, we can obtain $\kbinom(\beta,N,p)$ by looking at the intersection of the CDF with the horizontal line $y\equiv\beta$.
The key insight from the derivation of the above framework is that using the naïve SAA approach, \ie $\kthres=\eta N$, corresponds to a confidence $\beta\approx0.5$.
Indeed, in Fig.~\ref{fig:cdf_against_eta}, we can see that $\kbinom(0.5,N,p)\approx pN$ because the horizontal line $y\equiv0.5$ intersects the curves roughly at $k/N=p$.
This highlights a key limitation of the naïve approach: by implicitly operating at a confidence level of around 50\%, it tends to accept solutions that have a high chance of violating the true constraint, despite appearing feasible on the sampled data.
This is further illustrated in Sec.~\ref{sec:appendix_naive_vs_confidence} and Fig.~\ref{fig:cc_binomial_analysis} in the Appendix.

The idea of expressing a chance constraint in terms of a sample-based variable $s_N(\xb;D)$ and the inverse CDF, \ie $\mathrm{CDF}^{-1}(\beta)$ is not new; see, \eg \citet{heirung2018stochastic, pena2020solving}. In fact, it has been used in the past to derive theoretical bounds on the number of samples needed to ensure constraint satisfaction in scenario optimisation approaches; see, \eg \citet{campi2011sampling}. 
Yet, to the best of our knowledge, it has never been used within the Boolean formulation of a chance constraint and its sample average approximation. Instead, it has only been used for simple continuous constraints with tractable distributions for which the CDF could be derived analytically. 

\subsection{Limitation of the Approximation}\label{sec:limitation-bernoulli}
In summary, the confidence-bounded surrogate constraint Eq.~\eqref{eq:surrogate} is based on the formulation of the SAA as a Bernoulli process.
However, a crucial assumption in Proposition~\ref{prop:cc_binom} is that the Bernoulli variables $G_{\xb,i}=\boldsymbol{1}_{g(\xb,\deltab_i)>0}$ for $i=1,\ldots,N$, are independent, such that the sum of the Bernoulli variables $s_N(\xb;D)$ follows a binomial distribution.
However, when applied to the output of \eqref{eq:cc_opt_surrogate}, although the samples $\{\deltab_i\}_{i=1}^N$ themselves are independent, the Bernoulli variables $G_{\xb,i}$  are generally not independent, as the solution $\xb$ depends on these samples through the optimisation scheme. 
Hence, only if $\xb$ is independent from the samples $\{\deltab_i\}_{i=1}^N$, does $s_N(\xb;D)$ follow a binomial distribution, and Proposition~\ref{prop:cc_binom} holds.
Said otherwise, Proposition~\ref{prop:cc_binom} gives the probability of rejecting a \emph{given} infeasible solution $\xb_{\mathrm{reject}}$, independent of the samples.
Nevertheless, the probability of rejecting \emph{all} infeasible solutions is in general strictly smaller.
Consequently, without additional assumptions that ensure independence, solving the surrogate optimisation problem Eq.~\eqref{eq:cc_opt_surrogate} with the confidence-bounded threshold $\kbinom$ is a \emph{heuristic} approach.

\subsection{Addressing the Limitation}

In the following, we present two settings where the optimisation framework \eqref{eq:cc_opt_surrogate} can be applied under reasonable independence assumptions. These represent cases where our heuristic approach offers \emph{firm guarantees}.
For the second setting, this also provides insight into why the surrogate performs well in practice, even if the underlying assumption is difficult to verify in practice, or may be satisfied only part of the time.   

\subsubsection{A-posteriori validation}
In the first setting, Eq.~\eqref{eq:cc_opt_surrogate} is used as an \textit{a-posteriori} validation step. This works as follows: Given a candidate solution $\hat{\xb}$, \eg obtained by solving Eq.~\eqref{eq:cc_opt_surrogate} using a sample set $\hat{D}$, we re-evaluate the constraint function $g$ at $\hat{\xb}$ using a new set of $N$ i.i.d.~samples $D = \{\deltab_i\}_{i=1}^N$. If the empirical violation count $s_N(\hat{\xb}; D)$ exceeds the acceptance threshold $\kbinom$, the candidate solution is rejected.

\begin{proposition}[(Informal)]
If the candidate solution passes the validation test, then with confidence $1-\beta$ it satisfies the true chance constraint.
\end{proposition}

\begin{proof}
This setup falls within the scope of Proposition~\ref{prop:cc_binom}, since $\hat{\xb}$ and the validation set $D$ are independent. As a result, if $\hat{\xb}$ violates the true chance constraint, Proposition~\ref{prop:cc_binom} guarantees that it will be rejected with probability at least $1 - \beta$ during the \textit{a-posteriori} validation. Conversely, if the candidate passes validation, we can assert with confidence at least $1 - \beta$ that it satisfies the chance constraint.
\end{proof}

However, this raises the question of how to proceed when a candidate solution is rejected. One option is to repeat the optimisation and validation steps until a candidate passes. Yet, doing so introduces a \emph{multiple hypothesis testing problem}: although each individual test maintains a confidence level of $1 - \beta$, the \emph{overall} probability of accepting a violating solution increases unless this accumulation is properly corrected. Statistical correction methods, such as those discussed by~\citet{abdi2007bonferroni}, can mitigate this issue, though at the cost of increased conservatism or a higher required sample size. A more rigorous analysis of \textit{a-posteriori} validation as a verification mechanism for certifying the safety and performance of robotic policies is provided in~\citet{vincent2024guarantees}.

In time-sensitive applications such as online planning, performing multiple rounds of optimisation and validation, potentially with increasing sample counts, may not be feasible. In such cases, it is advisable to cap the number of candidate evaluations (or limit the computation time) and instead rely on fallback strategies or recovery controllers to ensure constraint satisfaction when no validated solution is found.

\subsubsection{Receding-horizon optimisation}
The second setting where our approach can offer firm guarantees is the receding-horizon MPC context, where Eq.~\eqref{eq:cc_opt_surrogate} is solved repeatedly at each control step.
We provide an assumption (Assumption~\ref{ass:smooth}) under which a form of independence of the Bernoulli variables holds.
While this assumption may not strictly hold at every step or be directly verifiable in practice, it provides a plausible theoretical justification for the empirical effectiveness of our approach in the MPC setting.
In other words, even though our method is heuristic in the general setting, this setting gives a context in which it behaves reliably and aligns with our experimental observations (see Sec.~\ref{sec:exp_online}). We now formalize this intuition with an assumption and proposition:

At each MPC step $m=1,\ldots,M$, we compute a new solution $\xb_m$ with a new sample set $D=\{\deltab_{m,i}\}_{i=1}^N$.
That solution is executed for a few milliseconds, until we re-sample and compute a new solution in the next MPC step.
In general, the solution $\xb_m$ is \emph{not very different} from the solution $\xb_{m-1}$ (if needed, this can also be enforced as an explicit constraint of the MPC).
In this case, provided the constraint function $g$ varies smoothly with $\xb$, we can make the assumption that the binary trials $\boldsymbol{1}_{g(\xb_m,\deltab_{m,i})>0}$ are independent because $\{\deltab_{m,i}\}_{i=1}^N$ is i.i.d.~and $\xb_m\approx\xb_{m-1}$.
We formalize this with an assumption and a proposition:

\begin{assumption}\label{ass:smooth}
There is $\epsilon>0$ such that with probability $1-\epsilon$ on $\deltab$, if there is $\xb\in X$ such that $g(\xb,\deltab)>0$, then for all $\xb\in X$, it holds that $g(\xb,\deltab)>0$.
\end{assumption}

\begin{figure}
    \centering
    \includegraphics[width=0.5\linewidth]{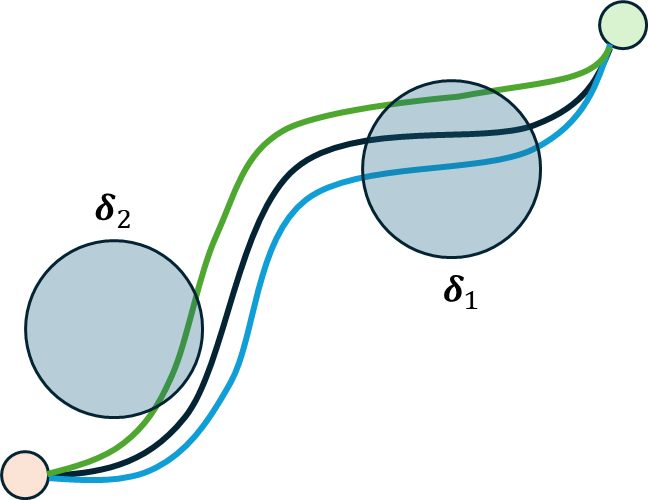}
    \caption{When $\deltab=\deltab_1$, the obstacle collides with all three paths, whereas when $\deltab=\deltab_2$, the obstacle collides only with the green path.
    Assumption~\ref{ass:smooth} states that $\deltab$ takes a value for which a situation like $\deltab_2$ occurs with probability at most $\epsilon$.}
    \label{fig:assumption-continuity}
\end{figure}

See Fig.~\ref{fig:assumption-continuity} for an illustration of Assumption~\ref{ass:smooth}.

\begin{proposition}\label{prop:cc_binom-smooth}
Under Assumption~\ref{ass:smooth}, it holds that
\begin{equation}\label{eq:proposition}
P_{\deltab_1\sim\Delta,\ldots,\deltab_N\sim\Delta}[P[G_{\xb}=1]\leq\eta+\epsilon] \geq 1-\beta,
\end{equation}
where $\xb$ is the solution of the surrogate optimisation problem in Eq.~\eqref{eq:cc_opt_surrogate} with $D=\{\deltab_i\}_{i=1}^N$ and $\kthres=\kbinom(\beta,N,\eta)$.
\end{proposition}

\begin{proof}
Let $\xb$ be the solution of Eq.~\eqref{eq:cc_opt_surrogate} with $D=\{\deltab_i\}_{i=1}^N$.
Assume that $P[G_{\xb}=1]>\eta+\epsilon$.
By Assumption~\ref{ass:smooth}, this implies that $P_{\deltab\sim\Delta} \left[ \min_{\xb'\in X} g(\xb',\deltab)>0 \right] > \eta$.
Furthermore,
\[
s_N^\txtmin(D)\triangleq\sum_{i=1}^N \min_{\xb'\in X} \boldsymbol{1}_{g(\xb',\deltab_i)>0} \leq s_N(\xb;D) \leq \kbinom.
\]
Since the variables $\min_{\xb'\in X} \boldsymbol{1}_{g(\xb',\deltab_i)>0}$ for $i=1,\ldots,N$, are i.i.d., it follows that $s_N^\txtmin(D)$ is a Bernoulli process with $N$ trials and $p=P_{\deltab\sim\Delta} \left[ \min_{\xb'\in X} g(\xb',\deltab_i)>0 \right]$.
Hence, the probability that $s_N^\txtmin(D)\leq\kbinom$ is equal to $C(\kbinom;N,p)$.
The rest follows in the same way as in the proof of Proposition~\ref{prop:cc_binom}.
\end{proof}


\begin{remark}
In a receding-horizon optimisation scheme, under the assumption that the solution does not vary too much from one step to the next one, the samples used at the previous steps also provide indication that the solution at the current step is valid.
More precisely, if the solution $\xb_m$ is equal (resp.~similar) to $\xb_{m-1},\ldots,\xb_{m-\ell+1}$, it implies that $\xb_m$ satisfies (approximately) the surrogate constraint $s_N(\xb_m,D_{m-k})\leq\kthres$ for all $0\leq k\leq\ell-1$, where $D_{m-k}=\{\deltab_{m-k,i}\}_{i=1}^N$.
Said otherwise, $\xb_m$ satisfies (approximately) the surrogate constraint with $\ell N$ samples, $s_{\ell N}(\xb_m,\bigcup_{k=0}^{\ell-1}D_{m-k})\leq\kthres$.
The larger number of samples in the surrogate constraint increases the confidence that $P[G_{\xb_m}=1]\leq\eta$.
However, this information is not used in the confidence bound that we can derive from Proposition~\ref{prop:cc_binom-smooth}, since it is for a single MPC step.
This is why in practice the real value of $P[G_{\xb_m}=1]$ is often smaller than $\eta$, as we will see in the experiments (see Sec.~\ref{sec:experiments}).
In future work, we plan to use and quantify this information to obtain even less conservative bounds on $P[G_{\xb_m}=1]$.
\end{remark}


\subsection{Relationship between Chance Constraints and Conditional Value-at-Risk}\label{sec:cvar}
In the following, we introduce the concept of \emph{conditional value-at-risk} (CVaR)~\citep{majumdar2020should} and its relationship to chance constraints in the context of binary indicator functions. This relationship will be helpful to understand subsequent results comparing the two formulations.
In contrast to formulating chance constraints, the concept of constraining the CVaR depends on the topology of the constraint function $g$ with respect to the realization of the disturbance. In general, \citet{lew2023risk} shows that using a chance constraint as in Eq.~\eqref{eq:cc_opt_basic} is equivalent to constraining the \textit{value-at-risk} (VaR) with
%
\begin{equation}
\label{eq:var}
    \mathrm{VaR}_\eta(g(\xb,\deltab)) = \inf_{\lambda \in \mathbb{R}} \left\{ \lambda \mid P_{\deltab\sim\Delta}[g(\xb,\deltab)>\lambda] \leq \eta \right\} \leq 0.
\end{equation}
Furthermore, it can be shown that constraining the $\mathrm{CVaR}$ is strictly more conservative, \ie Eq.~\eqref{eq:var} holds if $\mathrm{CVaR} \leq 0$ \citep{lew2023risk}.

In the following, we show that constraining the $\mathrm{CVaR}$ of a binary indicator function corresponds to a tighter chance constraint with a lower effective chance threshold. As the $\mathrm{CVaR}$ corresponds to the expected value of the constraint function for $g > \mathrm{VaR}$, the $\mathrm{CVaR}$ of a binary indicator function can be reformulated in terms of the probability of violating the constraint $g$, \ie
\begin{equation}
\label{eq:cvar}
    \mathrm{CVaR}_\eta(G_{\bm{x}}) = \begin{cases}
        \frac{P[G_{\xb}=1]}{\eta}, &\mathrm{if} P[G_{\xb}=1] < \eta, \\
        \hfill 1, &\mathrm{otherwise.}
    \end{cases}
\end{equation}
The $\mathrm{CVaR}_\eta$ of the binary indicator function in Eq.~\eqref{eq:cvar} is in the range $[0,1]$. Next, we may define a threshold $\mathrm{CVaR}_{\mathrm{max}} \in [0,1]$ in order to construct a constraint based on the $\mathrm{CVaR}$ with $\mathrm{CVaR}(G_{\bm{x}}) \leq \mathrm{CVaR}_{\mathrm{max}}$. By using Eq.~\eqref{eq:cvar}, it follows that constraining the $\mathrm{CVaR}$ yields another threshold on the probability of violating the constraint, \ie
\begin{equation}
\label{eq:cvar_eta}
    P[G_{\xb}=1] \leq \eta \, \mathrm{CVaR}_{\mathrm{max}} \Longleftrightarrow \mathrm{CVaR}_\eta(G_{\bm{x}}) \leq \mathrm{CVaR}_{\mathrm{max}}.
\end{equation}
Note that for any $\mathrm{CVaR}_{\mathrm{max}} \in [0,1)$, the resulting constraint is more conservative than the $\mathrm{VaR}$ constraint in Eq.~\eqref{eq:var}. 

This increased conservatism has practical implications in control and planning under uncertainty. In particular, CVaR-based constraints offer stronger safety guarantees by effectively lowering the allowable probability of constraint violation. However, this safety margin comes at the cost of increased conservatism, which may lead to overly cautious or suboptimal behavior in less risk-sensitive settings. Therefore, understanding the trade-off between chance constraints and CVaR constraints is essential for selecting the appropriate level of risk aversion based on the application's requirements, as we will show in Sec.~\ref{sec:experiments}.


\section{Stochastic Trajectory Optimisation with Chance Constraints}\label{sec:ccvpsto}

Due to the non-smooth nature of the uncertainty dynamics and the resulting non-smooth surrogate chance constraint with respect to the optimisation variable in Eq.~\eqref{eq:surrogate}, we approach the optimisation problem with a gradient-free, \ie zero-order, evolutionary optimisation technique. 
Building upon our previous work \emph{Via-Point-Based Stochastic Trajectory Optimisation (VP-STO)} \citep{jankowski2022vp}, we introduce \emph{chance-constrained VP-STO (CC-VPSTO)} for finding robot trajectories that minimise a given task-related objective \emph{while satisfying a given chance constraint}.

\subsection{Preliminaries on VP-STO}
VP-STO builds on stochastic optimisation in order to find robot trajectories that minimise a given task-related objective in dynamic environments. 

\subsubsection*{Trajectory Representation}
 In VP-STO the decision variable $\xb$ for an optimisation problem, such as the one in Eq.~\eqref{eq:cc_opt_basic}, is a set of $S$ via-points $\qb_{\mathrm{via}}=(q_{\mathrm{via},1},\ldots,q_{\mathrm{via},s})$, \ie $\xb=\qb_{\mathrm{via}}$. For a given set of via-points, VP-STO synthesises a time-continuous and smooth trajectory that satisfies the boundary conditions, such as initial and final state and velocity, and kinodynamic constraints, such as velocity and acceleration limits\footnote{For more information on how we generate the continuous trajectories from via-points, we refer the reader to Sec. \ref{sec:obf} in the Appendix and to \citep{jankowski2022vp, Jankowski2022}.}.
 The advantage of the approach lies in the low-dimensional representation of the trajectory, which allows for efficient optimisation in a low-dimensional space. 

\subsubsection*{Optimisation Algorithm} 
VP-STO uses the Covariance Matrix Adaptation Evolution Strategy (CMA-ES) \citep{Hansen2016} to find the optimal set of via-points that map to a trajectory that minimises the given objective function. CMA-ES iteratively updates the mean and covariance of a Gaussian distribution that represents the search space of the optimisation problem, \ie the set of via-points. In each iteration $j$, the algorithm samples candidate solutions from this distribution, \ie $\mathcal{N}\left({}^j\bm{\mu}_{\mathrm{via}}, {}^j\bm{\Sigma}_{\mathrm{via}}\right)$, evaluates them on the given objective function, and updates the distribution based on the evaluation results. The algorithm converges to the optimal solution in a few iterations, making it suitable for real-time applications. 

\subsection{Chance-Constrained VP-STO}
VP-STO has been shown to be effective in generating robot trajectories in real-time for dynamic environments, outperforming state-of-the-art sampling-based MPC methods~\citep{bhardwaj2022storm}. Yet, in its original form, VP-STO does not consider uncertainty, but instead assumes a deterministic environment. In this work, we extend the VP-STO framework to consider uncertainty in the environment, \ie we introduce the \emph{chance-constrained VP-STO (CC-VPSTO)} framework. 

CC-VPSTO solves the optimisation problem in~\eqref{eq:cc_opt_surrogate} using the surrogate constraint in Eq.~\eqref{eq:surrogate}. We enforce the constraint through a penalty-based approach, \ie we include the constraint $k\leq\kthres$ in the objective function as a penalty term. This can be seen as a discontinuous barrier function that adds a very high penalty term $J_{\mathrm{pen}}$ to the objective function if the constraint is violated, \ie when the observed number of constraint violations $k>k_{\mathrm{thresh}}$. The closed-form formulation of this penalty term is as follows: 
\[
J_{\mathrm{pen}} = \onebf[k > \kthres] \cdot (J_{\mathrm{pen, min}} + a\cdot(k - \kthres - 1)).
\]
We choose the minimum penalty term $J_{\mathrm{pen, min}}$ to be much larger than the maximum cost objective without constraint violations. Moreover, we add a piecewise linear term to the minimum penalty term that grows linearly with the extra number of violations compared to $\kthres$.
This term makes the constraint landscape smoother and gives the optimiser a direction towards feasible solutions without violations. We note that the specific magnitude of the penalty does not significantly influence CMA-ES behavior, provided it acts as a \emph{tight barrier} that clearly separates feasible from infeasible solutions. In this regime, CMA-ES reliably avoids constraint-violating regions, given that a feasible solution exists and the covariance is sufficiently large to explore the solution space. A key advantage of using stochastic, derivative-free optimization in this setting is that it allows us to employ an actual barrier-style penalty rather than smooth approximations such as log-barriers, which can introduce numerical sensitivity and tuning challenges.

\begin{figure}
    \centering
    \includegraphics[width=0.9\linewidth]{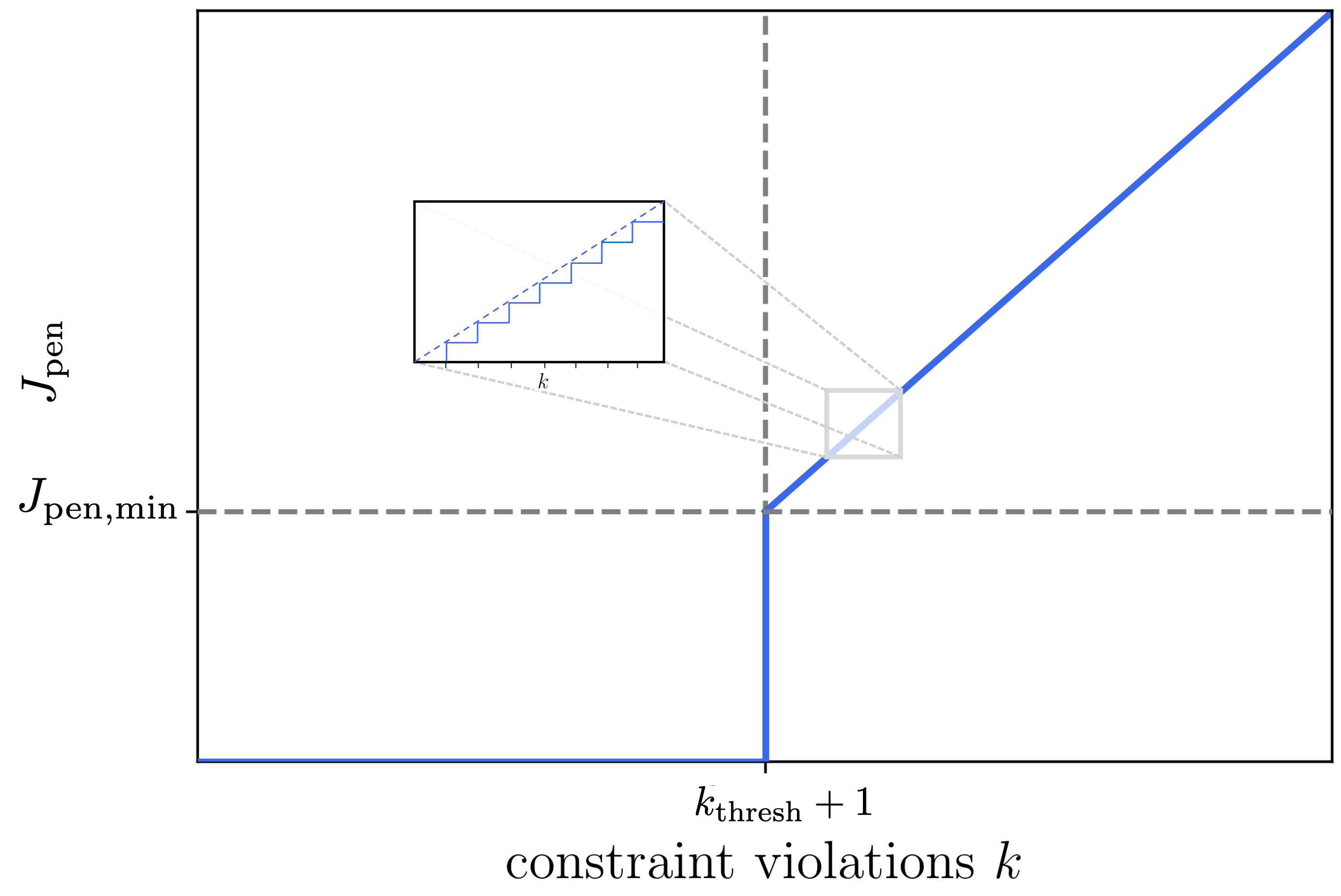}
    \caption{Graph of the penalty function used in CC-VPSTO.
    When observing more than $k_{\mathrm{thresh}}$ constraint violations in the $N$ Monte Carlo simulations, the penalty function takes value $J_{\mathrm{pen,min}}$ plus a quantity proportional to the number of extra constraint violations compared to $\kthres$.
    Note that we chose the minimum penalty term $J_{\mathrm{pen, min}}$ to be much larger than the largest cost objective without constraint violations.} 
    \label{fig:barrier_fct}
\end{figure}
The overall algorithm for CC-VPSTO is summarised in Algorithm~\ref{alg:ccvpsto}, where the approximation of the chance constraint, \ie counting the number of samples that cause the solution to violate the constraint, is encapsulated in the \texttt{evaluate} function.

In our previous work, we demonstrated the suitability of the VP-STO framework for real-time robot motion planning in dynamic environments~\citep{jankowski2022vp}. Similarly, the CC-VPSTO framework, \ie the above algorithm, can be used in a receding horizon MPC scheme to generate robot trajectories in real-time. Yet, we note that the constraint evaluation in the \texttt{evaluate} function will be computationally more expensive than in the original VP-STO framework, as it requires $N$ Monte Carlo simulations per candidate trajectory $\bm{\xi}$. This implies that the reactivity of our framework now depends on the number of samples $N$ used in the approximation. The advantage of our approximation is that it allows us to choose the number of samples and then use confidence levels to determine an appropriate threshold.
Crucially, the number of samples can be selected based on the target execution frequency of the MPC scheme, enabling a trade-off between computational efficiency and approximation accuracy.
In contrast, scenario-based optimisation typically requires a large number of samples, derived from conservative theoretical bounds that do not account for real-time constraints. This limits its practical applicability in time-sensitive settings.
Last, we note, that VP-STO and thus also CC-VPSTO in the current form does not handle non-holonomic constraints, such as those arising from differential drive robots, which we leave for future work.

\RestyleAlgo{ruled}
\newcommand{\mycomment}[1]{\Comment{\footnotesize\textcolor{blue}{#1}}}
\newcommand\mycommfont[1]{\scriptsize\ttfamily\textcolor{blue}{#1}}
\SetCommentSty{mycommfont}
\SetKwComment{Comment}{\footnotesize\textcolor{blue}{/* }}{\footnotesize\textcolor{blue}{ */}}
\SetKw{AND}{and}
\SetKw{IN}{Input:}
\CommentSty{\tiny\color{blue}}
\begin{algorithm}[t!]
    \small
    \DontPrintSemicolon
    \caption{CC-VPSTO}\label{alg:ccvpsto}
    \KwIn{$\qb_0, \dot{\qb}_0, \qb_T, \dot{\qb}_T, \dot{\qb}_{\mathrm{min}}, \dot{\qb}_{\mathrm{max}}, \ddot{\qb}_{\mathrm{min}}, \ddot{\qb}_{\mathrm{max}}$, $N_{via}$, $maxIter$, $S$, $H$, $\eta$, $\beta$, $N$}
    \mycomment{$N_{via}$: no. of via-points, \\ $maxIter$: max. no. of CMA-ES iterations, \\ $S$: no. of sampled candidate trajs.\\$H$: horizon\\$\eta$: chance constraint threshold\\$\beta$: confidence threshold\\$N$: no. of samples}
    \KwOut{Robot trajectory $\bm{\xi}_{0:H}^*$}
        ${}^0\bm{\mu}_{\mathrm{via}}, {}^0\bm{\Sigma}_{\mathrm{via}} \gets$ init($N_{via}$)\; 
        $j \gets 0$\;
        Sample $\Delta \gets \{\deltab_i \sim p_\Delta\}_{i=1}^N$\;
        $k_{\beta} \gets \kbinom(\beta,N,\eta)$\;
        \While{$ j < maxIter$}{
            $\{\qb_{\mathrm{via}}\}_{s=1}^S \gets$ sample$\left({}^j\bm{\mu}_{\mathrm{via}}, {}^j\bm{\Sigma}_{\mathrm{via}}\right)$ \tcp*{via-points}\;
            $\{\bm{\xi}\}_{s=1}^S \gets$ synthesise$\left(\{\qb_{\mathrm{via}}\}_{s=1}^S\right)$ \tcp*{trajectories}\;
            $\{c\}_{s=1}^S \gets$ evaluate$\left(\{\bm{\xi}\}_{s=1}^S, k_{\beta}, \Delta \right)$ \tcp*{cost}\; 
            ${}^{j+1}\bm{\mu}_{\mathrm{via}}, {}^{j+1}\bm{\Sigma}_{\mathrm{via}} \gets$ CMA-ES$\left(\{\qb_{\mathrm{via}}, c\}_{s=1}^S\right)$\;
            $j \gets j+1$\;
        }
        $\bm{\xi}^*_{0:H} \gets$ synthesise$\left(\bm{\mu}^{j}_{\mathrm{via}}\right)$\;
    \vskip -2pt
\end{algorithm}


\section{Experiments}\label{sec:experiments}

We evaluate our framework, \ie Algorithm~\ref{alg:ccvpsto}, with and without the MPC scheme, in simulations and in a real-world experiment with a Franka Emika robot arm.
The simulation experiments allow us to make claims about the empirical performance of our system across different settings and parameterisations. The robot experiment allows us to evaluate the real-time applicability of our approach. The supplementary video includes videos from both simulated and real experiments. These can also be found on our website \href{https://sites.google.com/oxfordrobotics.institute/cc-vpsto}{https://sites.google.com/oxfordrobotics.institute/cc-vpsto}.

\subsection{Experimental Setup}
\label{sec:exp_clarification}
\subsubsection{Joint probability of constraint violation} In all our experiments the chance constraint is formulated on the collision probability with obstacles in the robot's environment. We encode this as a \emph{joint} chance constraint, \ie enforcing \emph{trajectory-wise} constraint satisfaction with high probability (cf. Sec.~\ref{sec:related_work} for more details).
A joint formulation is more meaningful interpretation for robot behaviour, in contrast to evaluating the constraint independently at each time step. This means that we would not consider a trajectory to be safe if it avoids collisions in one time step with a very high probability, but collides in the next time step. We thus consider correlation over time in the chance constraint, \ie the first collision in a trajectory renders the whole trajectory unsafe and all subsequent collisions do not add any additional risk~\citep{lew2023risk}. 

\subsubsection{Uncertainty Samples} 
Before outlining our experiments in detail, we first clarify the role of uncertainty samples in both \textit{i)} our algorithm and \textit{ii)} its evaluation.
In each of the experiments, we draw separate sample sets for the optimisation and for reporting satisfaction of the chance constraint within the experiment. 
Across all experiments, an uncertainty sample corresponds to a single possible realisation of how the environment may evolve, in other words, a scenario.
a sample represents a specific obstacle position drawn from a distribution (\eg a Gaussian).
When multiple dynamic obstacles are present, a single sample consists of $M$ predicted trajectories, one for each of the $M$ obstacles.
The uncertainty distribution used for both optimisation and evaluation is assumed to be the same and fixed throughout the experiment. In the following, we consistently use a confidence level of $1-\beta = 0.95$ (\ie $\beta = 0.05$) across all experiments.

\subsubsection{Collisions}
In the case of a single obstacle, we consider a robot trajectory to be in collision if the robot collides with the obstacle at \textit{any point in time} across the whole trajectory. For multiple obstacles, we extend this definition to say a robot trajectory is in collision if the robot would collide with \textit{any} obstacle \textit{at any point in time}. By this, we avoid double counting collisions. One uncertainty sample can only be counted as one collision, even in cases where it might collide with several obstacles at different points in time.

\subsection{Simulation Experiments}\label{sec:simulation_experiments}

All simulation experiments are conducted in a bounded 2D environment with a circular holonomic robot, as shown in Fig.~\ref{fig:offline_analysis_results_1}. 
In Sec.~\ref{sec:exp_offline}, we perform offline planning experiments where CC-VPSTO is run once to compute a trajectory over the full horizon from start to goal under Gaussian uncertainty. In Sec.~\ref{sec:exp_offline_multi}, we evaluate the same offline setting but with multi-modal, non-Gaussian uncertainty to demonstrate the method's flexibility beyond standard distributional assumptions.
Last, in Sec.~\ref{sec:exp_online} we evaluate the online planning case, where we follow a receding horizon approach using CC-VPSTO to re-plan the trajectory at every MPC step.
The results of the offline experiments will also be relevant for the online setting, as each online replanning step can be seen as solving a new offline optimisation problem.
In both experiment settings, the uncertainty stems from obstacles in the environment, with no uncertainty in the robot dynamics\footnote{This is for simplicity only. The extension to process noise and external disturbances is straightforward given the proposed approach.}.
In the offline planning setting, obstacles are static but have uncertain positions, modeling the effect of measurement noise. In contrast, in the online receding-horizon setting, obstacles dynamically move according to a random walk model. Their velocities are reversed upon hitting workspace boundaries, keeping them within workspace bounds and introducing non-linear dynamics.
In all experiments, the obstacles are circular with varying radii, but our optimisation scheme does not rely on convexity and can accommodate more complex obstacle shapes.

\subsubsection{Offline Planning (Gaussian Uncertainty)}\label{sec:exp_offline}
In this offline planning setting, we show the properties of CC-VPSTO with a single static obstacle whose uncertain position follows a Gaussian distribution, as shown in Fig.~\ref{fig:offline_analysis_results_1} (see Sec.~\ref{sec:exp_offline_multi} for results on a non-Gaussian distribution). Here, a sample of the uncertainty refers to a possible (static) position of the obstacle, as explained in more detail in Sec.~\ref{sec:exp_clarification}.

For every combination of values of $N$ ($100$, $1000$), $\eta$ ($0.05$, $0.1$, $0.15$, $0.2$, $0.25$, $0.3$, $0.35$, $0.4$, $0.6$, $0.8$) and $\beta=0.05$, we run $N_{\mathrm{exp}}=10^5$ experiments. 
For $i=0,\ldots, N_{\mathrm{exp}}$ sample sets of $N$ i.i.d.\ samples, we compute the trajectory $\xi_i$ using CC-VPSTO with $\kbinom(\beta, \eta, N)$. 
Then, for each trajectory $\xi_i$, we evaluate its probability $\hat{\eta}_i$ of colliding with the static uncertain obstacle as follows. We use a new set of $N_{\mathrm{eval}}=10^4$ i.i.d.\ samples, sampled from the same distribution of possible obstacle locations, and count the number of samples that collide with $\xi_i$. 
The ratio of this number by $N_{\mathrm{eval}}$ is $\hat{\eta}_i$. 
We use $\hat{\eta}_i$ to compute the following three metrics:
\begin{enumerate}
    \item \textbf{\emph{Mean collision probability}:} \\ 
        $\hat{\eta}_{\mathrm{avg}}=\sum_{i=0}^{N_{\mathrm{exp}}} \hat{\eta}_i / N_{\mathrm{exp}}$
    \item \textbf{\emph{$(1-\bm{\beta})$-percentile of the collision probability}:} \\
        $\hat{\eta}_{(1-\beta)}=\mathrm{percentile}(\{\hat{\eta}_i\}_{i=0}^{N_{\mathrm{exp}}}, 1-\beta)$
    \item \textbf{\emph{Probability of chance constraint violation}:} \\
    $P(\hat{\eta}_i > \eta) = \hat{\beta} =  \Bigl(\sum_{i=0}^{N_{\mathrm{exp}}} \onebf_{\hat{\eta}_i > \eta} \Bigl)/ N_{\mathrm{exp}} $ 
\end{enumerate}
Note that we denote \emph{empirical} values with a hat, \eg $\hat{\eta}$.
%
%
For the proposed heuristic bound to be a good approximation, a proportion of maximum $\beta$ of the solutions can be in collision. This is because we set our confidence threshold to $1-\beta$
A trajectory $\xi_i$ violates the chance constraint if its estimated value $\hat{\eta}_i$ exceeds $\eta$.

\begin{figure}[t]
    \centering 
    \includegraphics[width=\columnwidth]{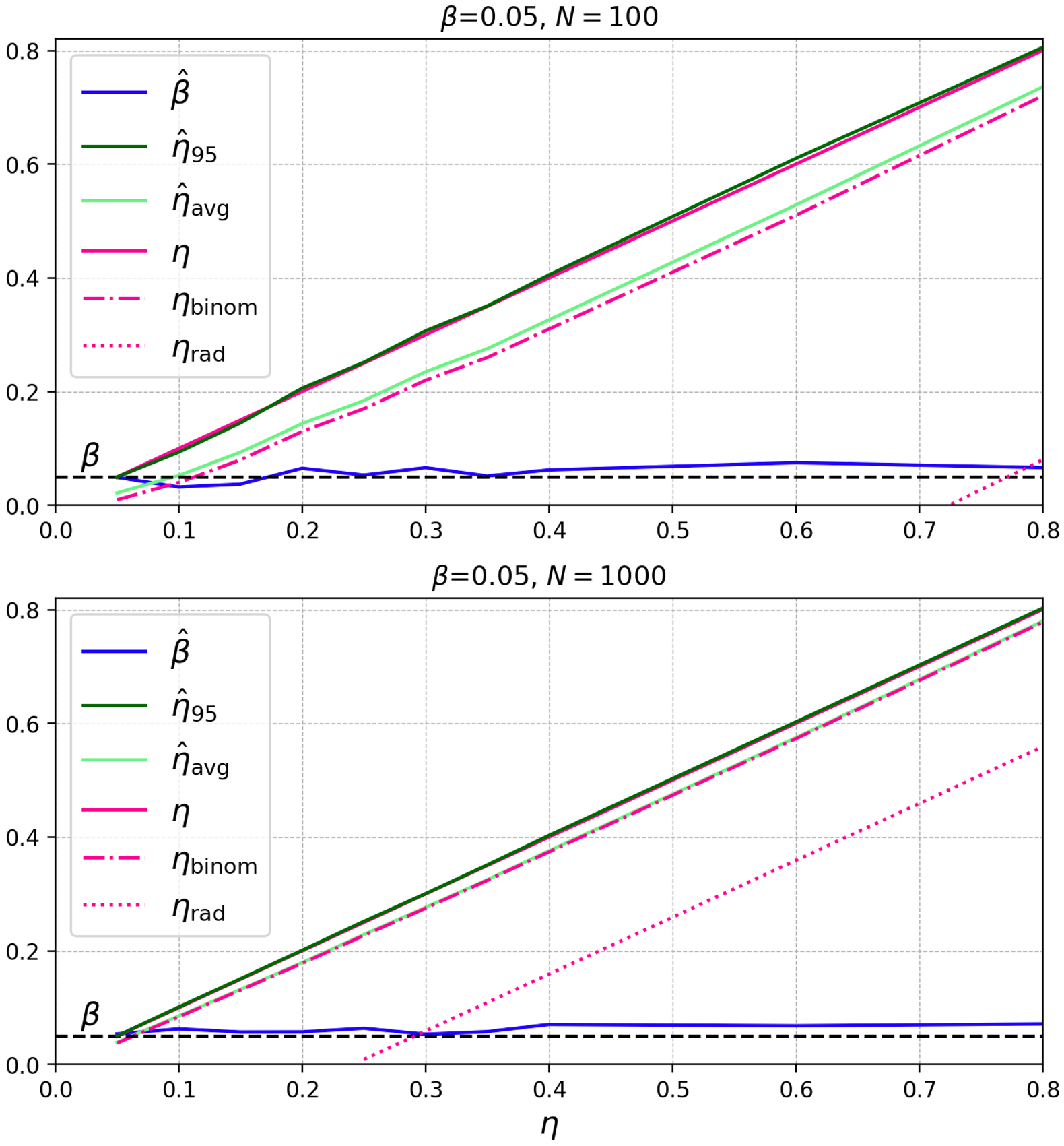}
    \caption{\emph{\textbf{Offline Planning Experiment.}}
    We evaluate the proposed binomial bound $\eta_{\mathrm{binom}}$ in a Gaussian offline-planning setting by running CC-VPSTO $N_{\mathrm{exp}}=10^5$ times for different risk levels $\eta$ and sample budgets $N\in\{100,1000\}$. Each resulting trajectory is then assessed on a new set of $N_{\mathrm{eval}}=10^4$ unseen obstacle samples to estimate its empirical collision probability $\hat{\eta}_i$. We report three aggregate metrics across experiments: the \emph{mean collision probability} $\hat{\eta}_{\mathrm{avg}}$, the \emph{$95$-percentile} $\hat{\eta}_{95}$, and the \emph{empirical chance-constraint violation rate} $\hat{\beta}$ (fraction of runs with $\hat{\eta}_i>\eta$). Theoretical curves for $\eta_{\mathrm{binom}}$ and the Rademacher-based baseline $\eta_{\mathrm{rad}}$ are shown for comparison.
    Importantly, the binomial bound $\eta_{\mathrm{binom}}$ (magenta, dash–dot) consistently provides a tight and accurate approximation of the true collision probabilities, especially for lower sample counts $N$.}
    \label{fig:offline_analysis_results}
\end{figure}
\subsubsection*{Baseline}
In the course of this work, we developed an alternative approach to approximate the chance constraint in Eq.~\eqref{eq:cc_opt_basic} based on the \emph{Rademacher complexity} from statistical learning theory \citep{shalev2014understanding,mohri2018foundations}. Computing a suitable $\kthres$ for the surrogate optimisation problem in Eq.~\eqref{eq:cc_opt_surrogate} can be approached by computing an upper bound on the Rademacher complexity of the associated set of functions. However, despite the theoretical attractivity of the this approach, computing an upper bound on the Rademacher complexity can be very challenging in general, and there is usually no closed-from expression for such bounds. Yet, we found a tight bound $\krad$ for a special case of collision-avoidance problem. This bound does not require the independence of the Bernoulli variables, but it is more conservative, computationally expensive, and, less general since it is limited to a specific motion planning problem. Consequently, we use this bound as a baseline for our simulation experiments. The full derivation, the closed-form expression for $\krad$, and the proofs and assumptions can be found in Appendix~\ref{sec:baseline}.

\subsubsection*{Results}
\begin{figure}[t]
    \centering
    \includegraphics[width=\columnwidth]{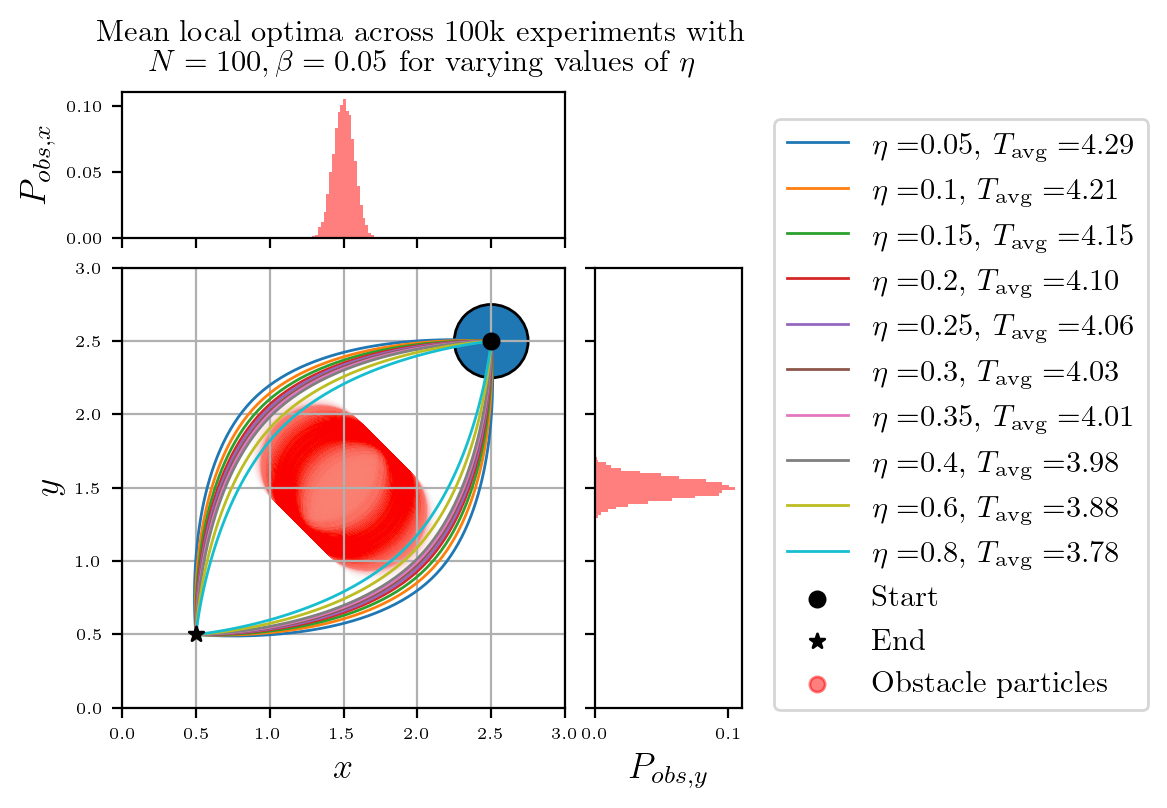}
    \caption{\emph{\textbf{Offline Planning Experiment (Gaussian Uncertainty).}} We show $N_{\mathrm{eval}}=10^4$ red circles for the uncertain obstacle position and the mean trajectory for the two local optima from CC-VPSTO, which used $N=100$ samples in the optimisation, for varying values of $\eta$ across $N_{\mathrm{exp}}=10^5$ experiments. The blue circle shows the robot's radius and starting position.}
    \label{fig:offline_analysis_results_1}
\end{figure}

The results of our offline analysis are summarised in Fig.~\ref{fig:offline_analysis_results}. The pink dotted curves show the theoretical values of $\eta_{\mathrm{binom}}$ and $\eta_{\mathrm{rad}}$, and the green curves show the empirical values of $\hat{\eta}_{\mathrm{avg}}$ and $\hat{\eta}_{(1-\beta)}$ for CC-VPSTO. In addition, we plot the empirical probability of chance constraint violation $\hat{\beta}$ in blue against the user-defined value of $\beta$. Note that we also provide the exact numerical results from Fig.~\ref{fig:offline_analysis_results} in Tab.~\ref{tab:offline_analysis_results} in the Appendix. 
We observe that our proposed bound $\kbinom$ provides a sufficient value for $\kthres$ since the observed $\hat\eta_{0.95}$ is close to the target $\eta$ (or equivalently, $\hat\beta$ is close to $\beta$). This means that empirically the probability of collision is below $\eta$, with confidence $95\,\%$.
We also observe that when $N$ is larger, $\eta_{\mathrm{binom}}$ and $\hat\eta_{\mathrm{avg}}$ are closer to $\eta$, implying that the surrogate optimisation problem becomes less conservative as the number of samples increases, given the same user-defined confidence-level.
This is expected since more samples provide a better approximation of the distribution; at the cost of increased computation time.
Last, the figure also shows that the baseline bound $\eta_{\mathrm{rad}}$ is much more conservative, as it is significantly smaller than $\eta_{\mathrm{binom}}$. Moreover, the offset from $\eta_{\mathrm{binom}}$ increases substantially when decreasing the number of samples $N$.
In addition to the quantitative results, we visualize the mean trajectories for the two local optima found by CC-VPSTO for different values of $\eta$ in Fig.~\ref{fig:offline_analysis_results_1}. Note, that we only show the solutions for the experiments with $N=100$ in the optimisation, as the solutions for $N=1000$ are visually indistinguishable. In the legend of Fig.~\ref{fig:offline_analysis_results_1}, we also show the average motion duration of the trajectories across experiments for the different values of $\eta$. This qualitative analysis shows that with higher values of $\eta$, CC-VPSTO finds more efficient, but also less conservative trajectories, as the mean trajectories are closer to the obstacle, since they allow for a higher probability of collision.  

\subsubsection{Offline Planning (Multimodal Uncertainty)}\label{sec:exp_offline_multi}
\begin{figure}[t]
    \centering
    \includegraphics[width=\columnwidth]{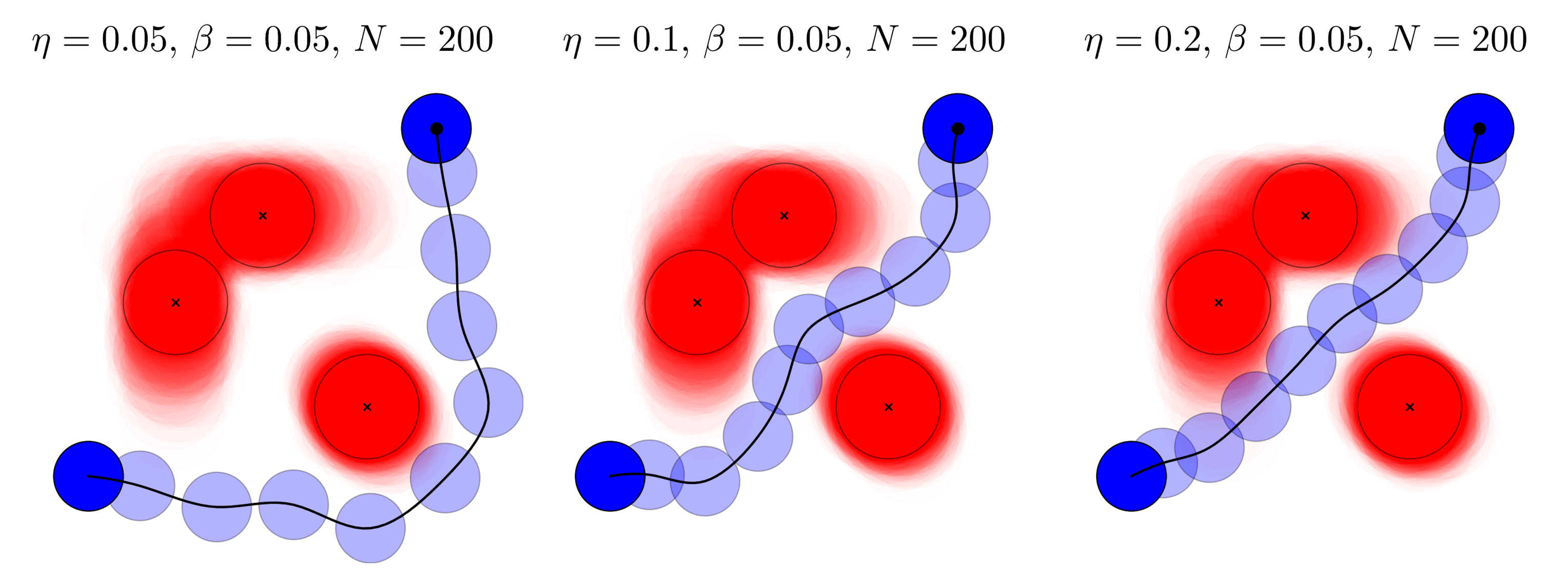}
    \caption{\emph{\textbf{Offline Planning Experiment (Multimodal Uncertainty).}} We show $N_{\mathrm{eval}}=10^4$ red circles for the uncertain obstacle position. The black crosses indicate the means of the three Gaussian modes. The black trajectory and the blue circles illustrate the optimal trajectory computed with CC-VPSTO, which used $N=200$ samples in the optimisation, for varying values of $\eta$.}
    \label{fig:offline_multimodal}
\end{figure}

CC-VPSTO does not make any assumptions about the uncertainty distribution and is able to handle arbitrary distributions. To illustrate this, we provide an additional offline motion planning experiment where we replace the Gaussian distribution over the obstacle position in the previous experiment from Sec.~\ref{sec:exp_offline} with a Gaussian mixture distribution with three modes. Fig.~\ref{fig:offline_multimodal} illustrates the qualitative results for this scenario with a multimodal distribution. It shows that CC-VPSTO is able to find a timing-optimal trajectory given the non-Gaussian uncertainty over the obstacle position depending on the user-defined chance threshold $\eta$ and confidence threshold $1-\beta$.
Table~\ref{tab:sim_res} provides numerical results on the average probability of collision of the solution ($\hat{\eta}_{\mathrm{avg}}$), the empiric estimation of $\beta$, and the average duration of the solution trajectory $T$ (i.e. the optimisation objective). These statistical results are computed from 1000 experiments performed for each $\eta$ and evaluated on $10^4$ new samples from the Gaussian mixture distribution. We observe that on average the chance constraint is satisfied. A higher chance threshold, \ie allowing a higher probability of colliding with the obstacle, results in lower trajectory durations.

\begin{table}[t]
\centering
\caption{\emph{Results of the Offline Planning Experiment (Multimodal Uncertainty)}.}
\renewcommand{\arraystretch}{1.1}
\label{tab:sim_res}
\begin{tabular}{lccc}
\toprule
& $\eta = 0.05$ & $\eta = 0.1$ & $\eta = 0.2$ \\ \midrule
\multicolumn{1}{l|}{$\eta_{0.95}$} & 0.038 & 0.084 & 0.178 \\
\hdashline

\multicolumn{1}{l|}{$\hat{\eta}_{\mathrm{avg}}$} & 0.026 & 0.086 & 0.164 \\
\multicolumn{1}{l|}{$\hat{\beta} \, (\beta = 0.05)$} &  0.026 & 0.0228 & 0.069 \\
\multicolumn{1}{l|}{$T_{\mathrm{avg}}$} & 4.702 & 3.164 & 2.752 \\ \bottomrule
\end{tabular}
\end{table}



A comparison of Table~\ref{tab:sim_res} and Figure~\ref{fig:offline_analysis_results} suggests that a multimodal distribution as the one given by the Gaussian mixture requires more samples in the approximation. This is to be expected because multi-modal distributions exhibit higher variance and can contain isolated high-probability regions that are easily missed with insufficient sampling. Capturing the shape and support of such distributions reliably in the sample-based surrogate requires a denser sampling of the space, which is why we used $N = 200$ samples in this experiment.

\begin{remark}
In view of the above, it is tempting to develop mechanisms for adapting $N$ to the uncertainty complexity (e.g., multimodality or high dimensionality). However, we note that in practice the complexity of the uncertainty is unknown to the planner and only accessible through samples, making this task non-trivial.
Therefore, we leave this direction for future work.
\end{remark}


\subsubsection{Online Planning (MPC)}\label{sec:exp_online}
The online planning experiments correspond to a receding horizon/model predictive control (MPC) approach, where a new robot trajectory is planned at every MPC step $t_{i, \text{MPC}}= t_{i-1, \text{MPC}} + \Delta_{\text{MPC}}$ with $1 / \Delta_{\text{MPC}}$ being the run frequency of the MPC controller.  
At each MPC step, the robot gets a position update of the $M$ obstacles in the environment, which is assumed to be exact, \ie no measurement uncertainty. As in the offline planning experiment, we assume that CC-VPSTO has access to a generative model that generates predictions of future obstacle motions, with samples that reflect the underlying uncertainty in those motions. In our online experiments we use a random walk model, parametrised to match the simulation environment. In a real-world setting, this could be replaced by a generative model learned from real-world data, \eg a model similar to \citet{jiang2023motiondiffuser}.  
In our optimisation scheme, given a position update, new obstacle trajectories are sampled from the random walk model and rolled out for a fixed time horizon $T>\Delta_{\text{MPC}}$. As described in Sec.~\ref{sec:exp_clarification}, one sample of the uncertainty in this experiment corresponds to one possible future of how the obstacles are going to evolve in the next time steps, \ie one sample maps to $M$ trajectory predictions of duration $T$ for $M$ obstacles.



\begin{figure*}[t!]
    \centering
    \includegraphics[width=0.95\linewidth]{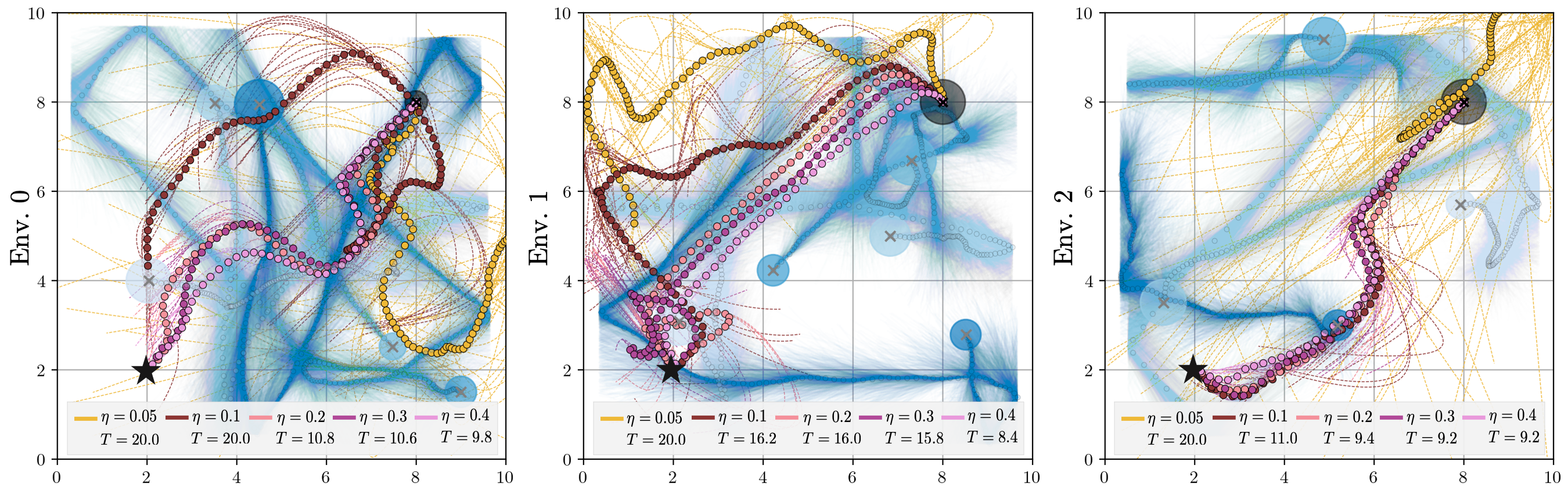}
    \caption{\emph{\textbf{Overview of the environments used in the MPC simulation experiments.}} 
     Each plot shows one example experiment setting for the respective environment configuration, along with the CC-VPSTO MPC solutions for varying $\eta$ values from a single experiment run. The initial obstacle positions and their radii are shown as blue circles. Smaller circles along the robot trajectory mark ground truth MPC updates, with the corresponding predicted sample rollouts visualized as semi-transparent trajectories originating from each update point. The robot’s start and goal are indicated by a dark grey circle (representing the robot radius) and a star, respectively. The current solution at each MPC step, \ie the trajectory segment planned over a receding horizon from the current robot position, is shown as a dashed line. Solutions get less conservative and more timing-effecient with growing values of $\eta$.
    }
  \label{fig:envs_overview}
\end{figure*}

\begin{figure*}[t]
    \centering
    \includegraphics[width=\textwidth]{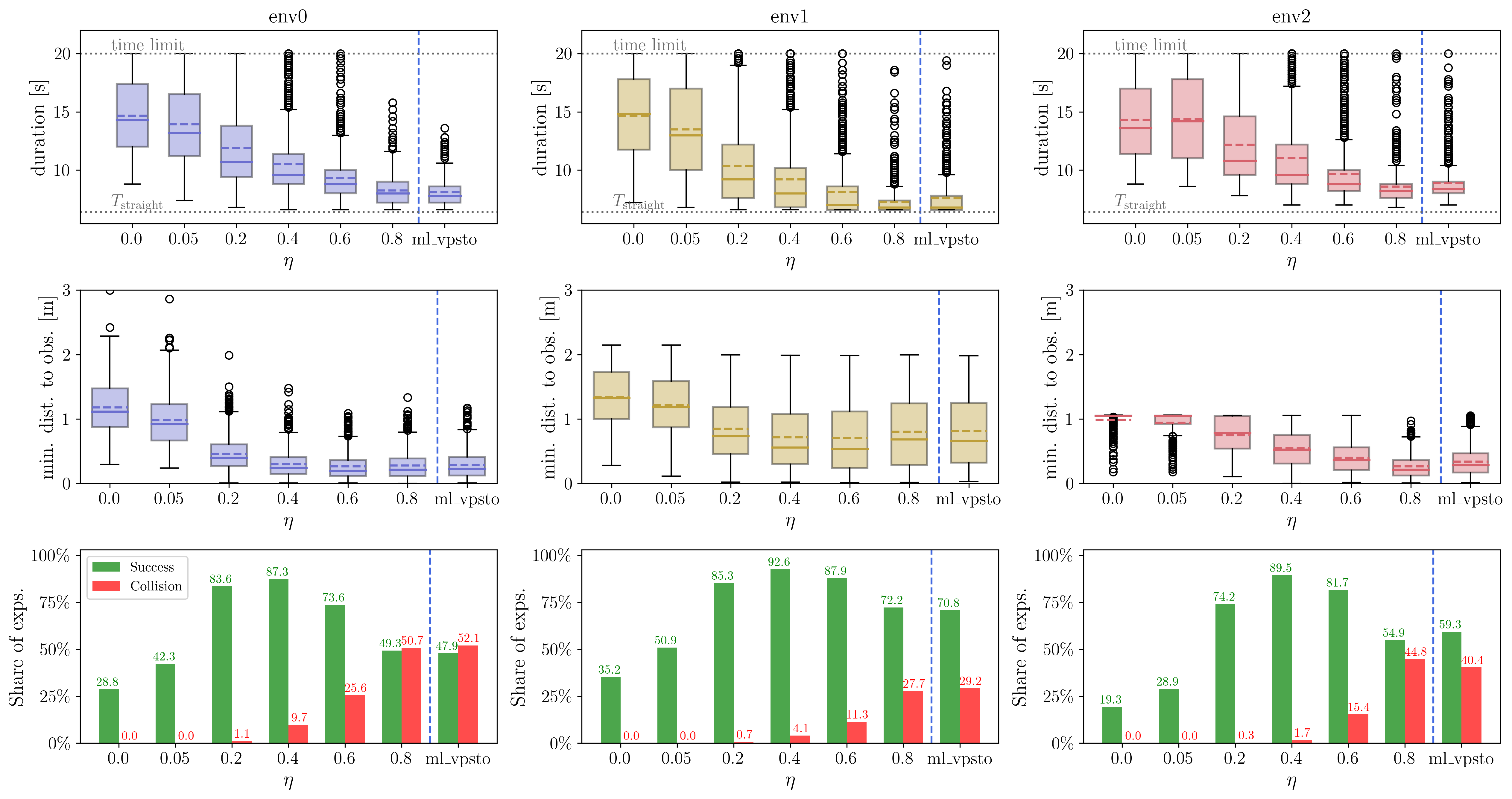}
    \caption{\textbf{\emph{Simulation: MPC experiments.}} Evaluating \emph{motion duration}, \emph{success rate}, \emph{collision rate} and the \emph{minimum distance to obstacles} across 1000 experiments on 3 different environments. One experiment corresponds to running online-CC-VPSTO until reaching the goal or until a maximum number of 100 MPC steps is reached. Goal and start location remain fixed across all experiments and environments, whilst the obstacle trajectories vary across experiments and environments. Each environment is initialized with different start positions and velocities of the obstacles, as well as different variance on the acceleration used in the random walk model. The boxplots include the mean (dashed line) and median (solid line) across all experiments.}
    \label{fig:mpc_experiments}
\end{figure*}


We evaluate online CC-VPSTO on four metrics across three environment configurations with four and five obstacles. Each obstacle is initialised with varying start position, velocity, and acceleration variance in the random walk model.
The trajectory we report results on is the trajectory that the robot executes, \ie the concatenation of the first $\Delta_{\mathrm{MPC}}$ time steps of each of the solutions across MPC steps. One single experiment produces one trajectory from the initial position to the goal for the given environment instantiation and $\eta$-value.
%
Fig.~\ref{fig:envs_overview} shows a qualitative example of each of the three environment configurations used in the MPC experiments, along with the respective CC-VPSTO solutions for different values of $\eta$. Note, that the length of the plotted obstacle trajectories was not fixed across the three examples, but depended on the maximum duration of the generated solutions for the given example. All solutions depicted are collision-free. Additional details about the environment configurations can be found in the Appendix in Sec.~\ref{sec:mpc_env_details}. For each environment configuration and for different values of $\eta$ (0.05, 0.2, 0.4, 0.6, 0.8), we run 1000 experiments.
%
The evaluation metrics for the experiments are:
\begin{enumerate}[leftmargin=*]
    \item \textbf{\textit{Motion duration (time until goal reached):} }This translates to the number of MPC steps needed until the robot reaches the goal. In the experiments, we set a maximum number of 100 MPC steps. We only report the duration for successful experiments.
    \item \textbf{\textit{Success rate:}} The fraction of experiments, where the generated trajectory reaches the goal within the maximum number of MPC steps. An experiment is further only considered to be successful if the executed robot trajectory does not collide at any point of time with any of the obstacles \textit{and} if the goal is reached.
    \item \textbf{\textit{Collision rate:}} This rate reflects the share of experiments where the respective trajectories were in collision at least once with any of the obstacles across the entire motion.
    \item \textbf{\textit{Minimum distance to obstacles:}} Per experiment, we measure the closest distance of the robot to any of the obstacles across the entire motion. This metric is only reported for successful experiments.
\end{enumerate}
Note, that these metrics are different from the metrics used in the offline evaluation. This is because we aim to show the properties of the MPC trajectory given the guarantees from the offline experiments (which corresponds to a single MPC step in the online case). 

\subsubsection*{Baselines}
We compare the use of our confidence-based bound $\kbinom$ as a surrogate for the chance constraint (cf. Eq.~\eqref{eq:surrogate-constraint}  and Eq.~\eqref{eq:surrogate}) against two alternative approximations: \textit{i)} the naïve MC approximation of the original chance constraint, as proposed in \eg \citet{blackmore2010probabilistic}, and \textit{ii)} the CVaR-based formulation described in Sec.~\ref{sec:cvar}, following the approach of~\citet{yin2023risk}. As the main contribution of this work lies in the derivation a new confidence-bounded sample average approximation of a chance constraint, we do not compare against other methods of solving the resulting optimisation problem, such as Model Predictive Path Integral Control (MPPI)~\citep{williams2017information}. For a more thorough comparison of VP-STO to MPPI, please refer to our previous work \citep{jankowski2022vp}.
In addition, we compare our approach to a baseline, that we abbreviate with ``ML-VPSTO'', where ML stands for maximum likelihood. Instead of computing the probability of constraint violation based on samples, ML-VPSTO uses the same samples to compute mean obstacle trajectories and uses standard VP-STO to generate a solution that avoids these trajectories. In addition, running CC-VPSTO with $\eta=0$ can also be seen as a baseline, as this is comparable to using a hard collision avoidance constraint within VP-STO. 
For all MPC simulation experiments, we assumed a replanning frequency of 4 Hz with a time step of 0.05 seconds, while setting the planning horizon $T_{\text{MPC}}$ to 5 seconds (mapping to 100 time steps for the rollouts), the maximum number of MPC iterations to 100 and the number of samples $N$ to 100.

\subsubsection*{Results}
The results of the online experiment are a key insight of this paper, as they demonstrate the effects of combining \emph{reactivity} (the MPC setting) with \emph{probabilistic bounds on constraint satisfaction} (the chance constraints). Fig.~\ref{fig:mpc_experiments} summarises the results across the 1000 experiments for each of the three different environment configurations. 
We observe that CC-VPSTO in an MPC loop is able to generate trajectories that are entirely collision-free for $\eta$ values of up to 5\%.  
In the given experimental setting, ML-VPSTO is approximately equivalent to permitting collisions with a probability as high as 80\% in CC-VPSTO. In $\texttt{env0}$, \ie the most challenging environment configuration, both ML-VPSTO and CC-VPSTO with $\eta=0.8$ lead to a situation where 50\% of the experiments are in collision. This indicates that employing average obstacle prediction for collision avoidance is inadequate, as a 50\% collision rate is generally not an acceptable outcome in the majority of robotic applications.
Moreover, the dependency of the constraint satisfaction/performance trade-off on the value of $\eta$ is reflected in the motion duration. The higher the value of $\eta$, the shorter the duration of the trajectory. While ML-VPSTO produces the quickest trajectories, it is also the least safe approach. For reference, we also include $T_{\text{straight}}$ in the duration plots, which is the duration of the straight-line trajectory from start to goal (ignoring obstacles).
When looking at the minimum distance to obstacles across trajectories and experiments, the expressiveness of this metric depends on the environment configuration. For the first and second environment configuration, there is a decreasing trend, until it plateaus at values of $\eta>0.4$ for the first environment. For the second environment configuration, after a downward trend, the minimum distance to obstacles increases again for values of $\eta>0.4$. This can be explained by CC-VPSTO being more risk-taking and probably choosing a more direct path to the goal, which can possibly lead to more collisions, but in the case of no collision, the distance to obstacles might actually be bigger, as the motion is also quicker and some obstacles might not have had time to move closer to the robot. Last, the small variance in the distances for small $\eta$ values in the third environment can be explained by the initial configuration of obstacles, as they are already very close to the robot, which is then probably already the minimum distance across the entire motion. 
Overall, for this experiment setting, it seems like CC-VPSTO with $\eta=0.4$ offers a good trade-off between constraint satisfaction and performance, as it is able to generate trajectories with a high success rate, whilst also being able to generate trajectories that are efficient in their motion duration. 

\begin{figure}
    \centering 
    \includegraphics[width=\linewidth]{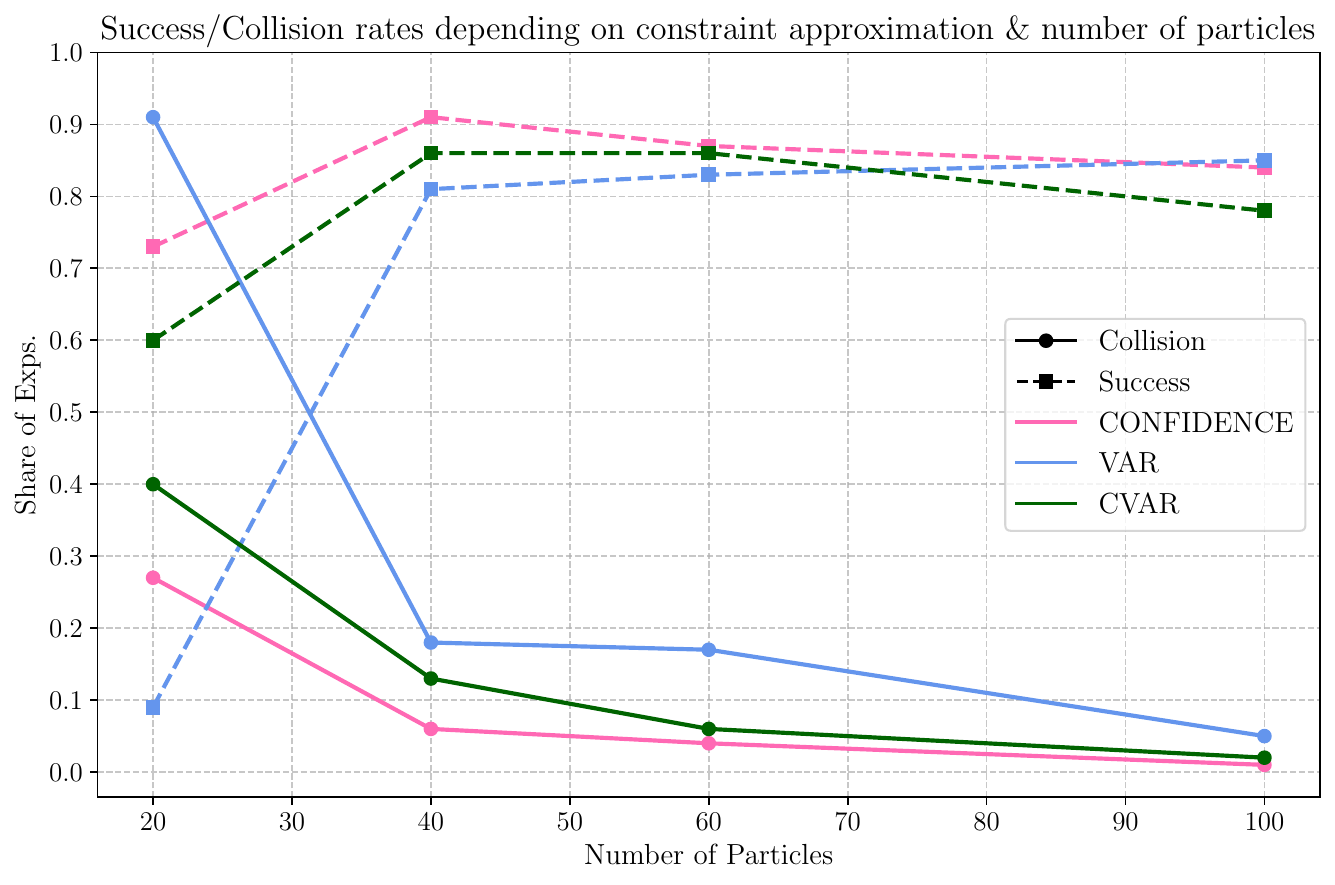}
    \caption{Comparison of the confidence-bounded chance constraint approximation to CVaR and a standard VaR approximation in terms of success and collision rates across MPC experiments depending on the number of samples used in the MC approximation.}
    \label{fig:constraint-type-vs-samples}
\end{figure}

Last, we evaluate the effect of the number of samples used in the MC approximation on the success and collision rates depending on the surrogate constraint that is used in CC-VPSTO. We do this across 1000 MPC experiments for each sample count and environment configuration. We evaluate three different surrogate constraints: 
\begin{enumerate}[leftmargin=*]
    \item \textit{Confidence:} $s_N(\xb,D)\leq\eta_{\text{binom}}(\eta,\beta)$
    \item \textit{Value at Risk (VaR):} $s_N(\xb,D)\leq\eta$
    \item \textit{Conditional VaR (CVaR):} $s_N(\xb,D)\leq \text{CVaR}_{\text{max}}$, with $\text{CVaR}_{\text{max}}$ set to 0.6. 
\end{enumerate}
For all experiments we fixed $\eta$ to 0.2 and the confidence threshold to 0.99. The results are shown in Fig.~\ref{fig:constraint-type-vs-samples}. The numbers are averaged across three different environment configurations. For 100 samples, the experiments are equivalent to the results shown in Fig.~\ref{fig:mpc_experiments}. With more samples, the collision rate converges to zero due to the MPC setting.
The results show that the confidence-bounded approximation is the only approximation that is able to maintain high success and low collision rates with a small number of samples.
This is in contrast to the other methods which do not account for the number of samples used in the approximation. With an increasing number of samples, VaR and our approximation converge to the same success and collision rates, which is expected as the number of samples increases, as shown in Sec.~\ref{sec:cvar}. Moreover, we notice that the success rate of CVaR decreases as the number of samples increases. Since CC-VPSTO explicitly accounts for the number of samples used in the approximation, its performance is less sensitive to the choice of $N$, showing the lowest variation across sample counts among the evaluated methods.
That said, in low-sample regimes, some approximation error remains unavoidable.

\subsection{Robot Experiment}\label{sec:robot_experiment}

\begin{figure}
    \centering
    \includegraphics[width=0.9\linewidth]{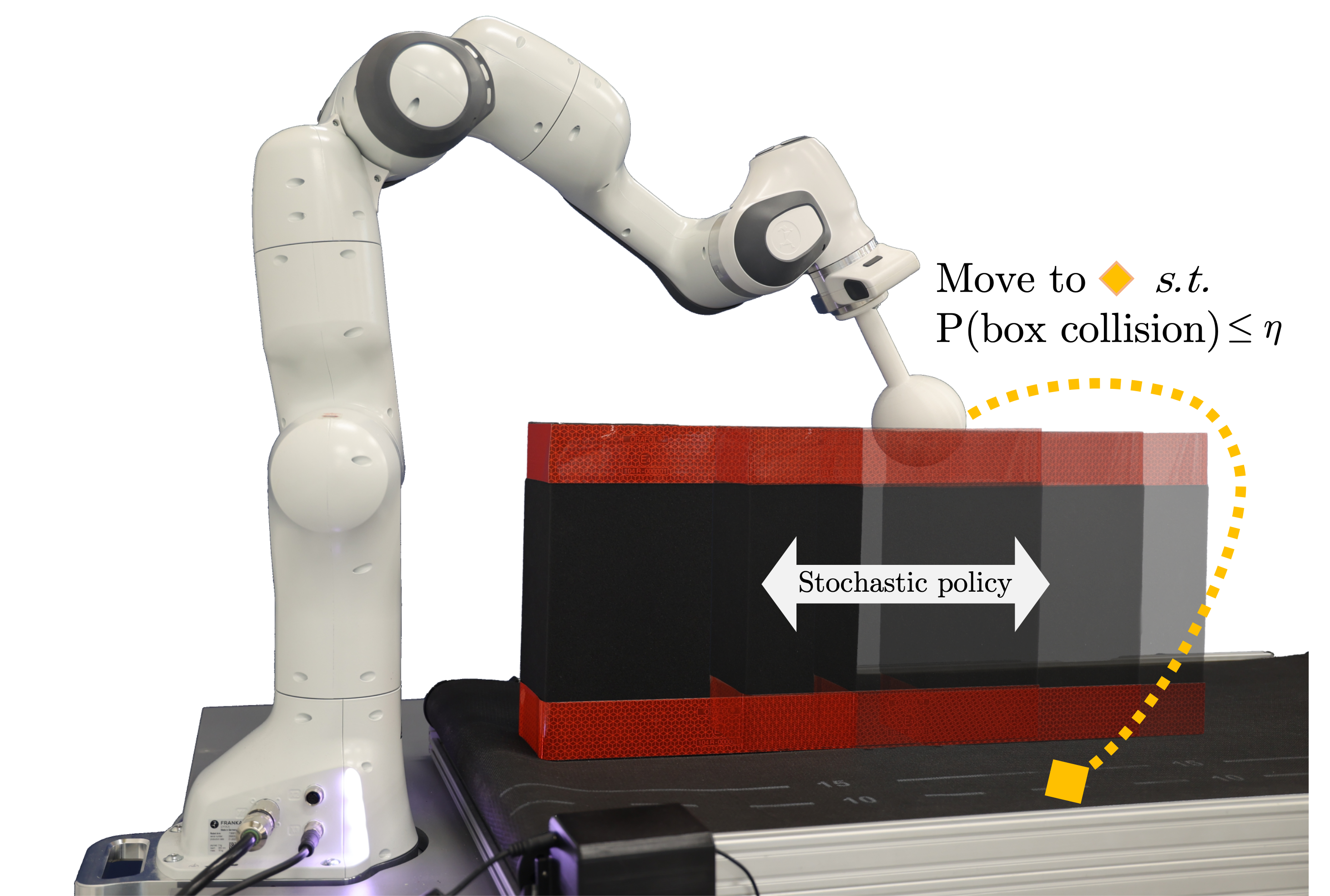}
    \caption{\textbf{\emph{Robot experiment setup.}} The robot task is to move from one side to the other side of the conveyor belt while assuring that the probability of colliding with the box obstacle is below a user-defined threshold $\eta$. 
    The motion of the box obstacle is stochastic, as the conveyor belt is actuated with constant velocity, but the direction of motion can change randomly at each time step with a probability that decays geometrically since the last direction change (see Appendix for details).
    }
    \label{fig:robot_exp_setup}
\end{figure}

We further demonstrate CC-VPSTO on a real robot for the scenario shown in Fig.~\ref{fig:robot_exp_setup}. The robot is tasked to move from one side to the other of the conveyor belt, whilst avoiding a box which is controlled according to a stochastic policy. This requires online, reactive motion generation that balances constraint satisfaction with task efficiency. The possible motions, also illustrated in Fig.~\ref{fig:robot_exp_teaser}, are to either move behind or in front of the box, as the robot is not allowed to simply move over the box. Moreover, besides the candidate trajectories that we synthesise from the sampled via-points in CC-VPSTO, we add a ``waiting'' trajectory to the set of candidate trajectories sampled in the final optimisation step. A waiting trajectory is a trajectory repeating the current robot position for the entire planning horizon, \ie keeping the robot stationary.  
This is to allow the robot to wait for the box to pass, which is a safe but not very efficient solution. Without these waiting trajectories, CC-VPSTO would keep the robot moving at all times, but this is not always necessary.

\vspace*{1ex}
\noindent\textbf{Setup.}
The experiment is performed on a Franka Emika robot arm. The framework was run on Ubuntu 20.04 with an Intel Core i7-8700 CPU@3.2GHz and 16GB of RAM. The ground truth box position is tracked using an Intel RealSense camera and a barcode detection pipeline. In every MPC step the robot is given the current position of the box and then plans a new trajectory using CC-VPSTO\footnote{Note we ignore measurement noise in this setup.}. With this setup, we are able to run the framework at a frequency of 3 Hz, using $N=100$ samples and a planning horizon of $T_{\text{MPC}}=3$ seconds (mapping to 60 time steps for the rollouts, as we use a time step of 0.05 seconds).

\vspace*{1ex}
\noindent\textbf{Stochastic conveyor belt policy.}
In this experiment, the uncertainty stems from the movement of the box on the conveyor belt, which serves as an obstacle for a robot to navigate around. The conveyor belt is velocity controlled, where the magnitude of the velocity is fixed to $0.05 \frac{\mathrm{m}}{\mathrm{sec}}$ but its direction is governed by a known probability density function. We describe our implementation of this stochastic model in more detail in the Appendix in Sec.~\ref{sec:robot_exp_details}.


\subsubsection*{Results}
\begin{figure}
    \centering
    \includegraphics[width=\columnwidth]{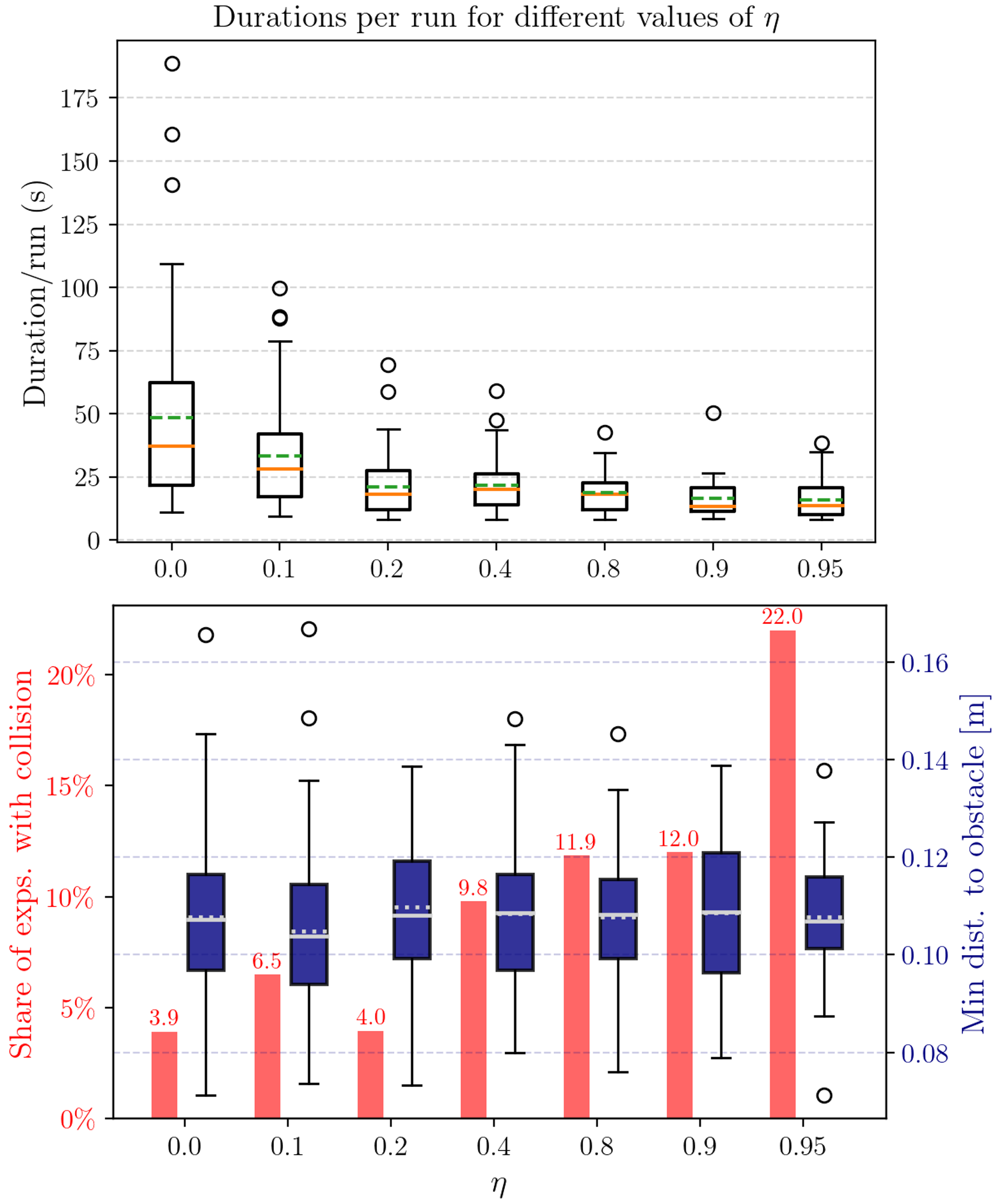}
    \caption{\textbf{\emph{Robot experiment results}.} Similar to the metrics evaluated in the MPC simulation experiments, we evaluate the \emph{motion duration}, \emph{minimum distance} to the box and the \emph{share of experiments that collide} with the box across 70 experiments for different values of $\eta$. Means in the boxplots are shown as dashed lines and medians as solid lines.}
    \label{fig:robot_exp_qualitative_result}
\end{figure}

Similar to the MPC experiments in simulation, we evaluate our real-world robot experiment on \textit{i)} the motion duration per run, \ie the time taken to go from one side of the conveyor belt to the other side, \textit{ii)} the share of experiments that collide with the box, and \textit{iii)} the minimum distance to the box across 70 runs for different values of $\eta$, as shown in Fig.~\ref{fig:robot_exp_qualitative_result}. 
We do not compare our approach to ``ML-VPSTO'' as for the given stochastic model, the mean is not useful. However, we include $\eta=0$ as a baseline, which corresponds to a VPSTO approach with a hard collision avoidance constraint. 
Overall, the results support the insights gained from the simulation experiments. 
We observe that the higher the value of $\eta$, the shorter the duration of the trajectory. This is because higher values of $\eta$ allow CC-VPSTO to generate trajectories that are more efficient, but also less safe.
Moreover, we also observe the trend that higher values of $\eta$ indeed lead to a higher share of experiments that collide with the box. However, we also observe that the share of experiments that collide with the box is still very low, even for very high values of $\eta$. This is an interesting insight from combining MPC with chance-constrained trajectory optimisation. In addition, we observe in this experiment that a value of $\eta=0.2$ outperforms $\eta=0.1$ in terms of the share of experiments that collide with the box. We anticipate that additional experiments will reduce this variability, as the current variance in the results is still quite high.
%
Last, in terms of the minimum distance to the box, we cannot observe a clear trend across different values of $\eta$. This can be explained by the use of waiting  trajectories that do not move the robot at all. The robot can choose to wait very close to the box, which is a constraint-satisfying but not very efficient solution. This results in higher durations, but not higher minimum distances.

\section{Discussion and Future Work}\label{sec:discussion}

Our experiments show that CC-VPSTO is able to generate task-efficient motions while consistently bounding the probability of constraint violation, \eg collision with stochastic obstacles. 
While it is typically more challenging to deal with chance constraints over entire trajectories (as opposed to constraints per time step), our SAA formulation allows us to do this in a straightforward way by simulating trajectories of the obstacles and the robot then checking for collisions between them at any time over a given horizon.
This is made possible by the flexibility of our approach which makes no assumption on the distribution of the uncertainty, but only requires sampling access to it. Hence, this can also be a joint distribution across all sources of uncertainty, \eg several obstacles. 

Consistent with our theoretical insights, the collision rate in the experiments remained below the specified threshold $\eta$, despite the fact that the independence assumption does not strictly hold in the receding-horizon (MPC) setting.
Nevertheless, we observe a gap between the collision rate and the threshold $\eta$.
This gap is much bigger in the online planning (MPC) than in the offline planning experiments.
This can be explained by the fact that at each MPC step we optimise trajectories for a longer horizon than just the time steps that we actually execute on the robot.
Hence, the anticipation of potential collisions in the future makes us more conservative, resulting in lower actual rate of collision than that which was imposed.
In future work, we plan to use discounted probabilities in the chance constraint, such as in \citet{yan2018stochastic}, to allow for larger probabilities of collision for time steps far in the future, knowing that the control input that we will apply then will be recomputed with stronger constraints in the meantime. Besides the introduction of discounted chance constraints, another direction could be to explore how to adapt the parameters $\eta$ and $\beta$ during online execution, \eg based on the current state of the system, the current uncertainty distribution, or the current cost function. This could result in a more robust and adaptive approach, which would be more suitable for real-world applications.

An important direction for future research is how to respond when a constraint violation actually occurs, especially since our framework allows such violations with some probability. While we do not propose an explicit or general mechanism for handling these situations, we assume that the robot is capable of recovering from violations. In this context, the MPC framework provides implicit robustness through continual replanning based on updated information about the uncertain environment. Nonetheless, it remains an open and valuable question how the planner's objective might be adapted in the face of violations; for instance, by temporarily shifting the objective to minimize the probability of further constraint violations rather than pursuing the original task goal.

Moreover, our work does not ensure recursive feasibility, which enforces that there always exists a solution to the optimisation problem.
Other works such as \citet{kohler2023stochastic}, have addressed this issue by enforcing that the predicted nominal state lands in a terminal set while not taking the measured state into account. 
However, this again comes at the cost of being restricted to linear systems with linear inequality constraints and convex objectives. For chance constraints this means that they would need to be linearized into half-space constraints, yet if we took the collision avoidance examples from this work and view them as a system that is augmented by the obstacle states, the non-collision constraint itself is non-linear in the augmented state as it has a quadratic relation through the distance measure. Yet, an interesting direction for future research would be how to ensure recursive feasibility through a less constrained problem formulation. 

A core assumption in our approach is that we are given a \emph{representative} model of the uncertainty, from which we can take samples. Yet, this might be a limitation in practice, as our model might not capture the true underlying distribution with sufficient accuracy, relating to \emph{epistemic uncertainty}. However, that can be addressed in future work, by either extending the approach to be \emph{distributionally robust}, such as in \citet{hakobyan2021wasserstein}, or by using generative data-driven models that can be adapted online as the robot acquires more data, similar to the work of~\citet{thorpe2022data}. In addition, future work should also quantify the extent to which MPC is able to provide inherent robustness, given that its closed-loop formulation allows for partial compensation when the true uncertainty distribution diverges moderately from the one used during optimisation.

In terms of computational efficiency, we have demonstrated the applicability of our algorithm to an MPC setting, where we achieve frequencies of 3 Hz on a real robot using 100 samples. The biggest computational bottleneck lies in the rollouts of the uncertainty dynamics. We therefore believe that the reported control rate can be improved by adding parallelization and GPU acceleration, which we did not leverage in the given experiments. However, multimodal distributions require more samples in the approximation, which might further limit current control rates.
We believe that future work could explore methods to efficiently generate representative sample sets, possibly leveraging learned generative models.
The advantage of our formulation is that the user can actively choose the number of samples while considering computational resources and requirements of minimum control rates. We also believe that there is still room for improvement in our implementation, as the sample rollouts for the stochastic box model have not been parallelised, as was done in the simulation experiments. Last, a learned generative model could also further improve the computational efficiency of the rollouts over the current Monte Carlo simulations, depending on the model's inference speed.




\section{Conclusion}
\label{sec:conclusion}
In this work, we addressed the problem of robot motion planning under uncertainty, aiming for both efficiency and constraint satisfaction in stochastic control settings.
We introduced a novel surrogate formulation for chance-constrained optimisation that enables statistically sound sampling-based motion planning under uncertainty. This, in turn, supports integration into a Model Predictive Control (MPC) framework for \textit{online}, reactive robot control.
The strength of our approach lies in its generality, as it does not require any specific assumptions on the underlying uncertainty distribution, the dynamics of the system, the cost function or the specific form of inequality constraints. 
While we focused on the problem of collision avoidance in this work, our approach is not limited to this problem, as it can be applied to any type of stochastic control problem, as long as we can sample from the uncertainty distribution. For instance, in future work we aim to extend this framework to include constraints on interaction forces in the context of contact-rich manipulation tasks and physical human-robot interaction. 
%
%
We showed that our approach is able to generate efficient trajectories that satisfy probabilistic constraints with high confidence across a variety of scenarios, including a real-world robot experiment.


\bibliographystyle{SageH}
\bibliography{refs}


\clearpage
\section*{APPENDIX}

\section{Trajectory Representation}\label{sec:obf}
The way we represent trajectories is based on previous work showing that the closed-form solution to the following optimisation problem
\begin{align}
\begin{split}
\label{eq:obf}
  &\textrm{min} \quad \int_0^{1} \bm{q}''(s)^\trsp \bm{q}''(s) ds \\
  \mathrm{s.t.} \quad &\bm{q}(s_n) = \bm{q}_n, \quad n = 1, ...,N\\
  &\bm{q}(0) = \bm{q}_0, \bm{q}'(0) = \bm{q}'_0, \; \bm{q}(1) = \bm{q}_T, \bm{q}'(1) = \bm{q}'_T
\end{split}
\end{align}
is given by cubic splines \citep{Zhang97} and that it can be formulated as a weighted superposition of basis functions \citep{Jankowski2022}. 
Hence, the robot's configuration is defined as $\bm{q}(s) = \bm{\Phi}(s) \bm{w} \in \mathbb{R}^D$, with $D$ being the number of degrees of freedom.
The matrix $\bm{\Phi}(s)$ contains the basis functions which are weighted by the vector $\bm{w}$\footnote{A more detailed explanation of the basis functions and their derivation can be found in the appendix of \citet{Jankowski2022}.}. 
The trajectory is defined on the interval $\mathcal{S} =  [0,1]$, while the time $t$ maps to the phase variable $s = \frac{t}{T} \in \mathcal{S}$ with $T$ being the total duration of the trajectory.
Consequently, joint velocities and accelerations along the trajectory are given by $\dot{\bm{q}}(s) = \frac{1}{T} \bm{\Phi}'(s) \bm{w}$ and $\ddot{\bm{q}}(s) = \frac{1}{T^2} \bm{\Phi}''(s) \bm{w}$, respectively\footnote{We use the notation $f'(s)$ for derivatives w.r.t. $s$ and the notation $\dot{f}(s)$ for derivatives w.r.t. $t$.}.
The basis function weights $\bm{w}$ include the trajectory constraints consisting of the boundary condition parameters $\bm{w}_{\textrm{bc}} = \left[ \bm{q}_0^\trsp, \bm{q}'^\trsp_0, \bm{q}_T^\trsp, \bm{q}'^\trsp_T \right]^\trsp$ and $N$ via-points the trajectory has to pass through $\bm{q}_{\textrm{via}} = \left[ \bm{q}_1^\trsp, \hdots, \bm{q}_N^\trsp \right]^\trsp \in \mathbb{R}^{D N}$, such that $\bm{w} = \left[ \bm{q}_{\textrm{via}}^\trsp, \bm{w}_{\textrm{bc}}^\trsp \right]^\trsp$. 
Throughout this paper, the via-point timings $s_n$ are assumed to be uniformly distributed in $\mathcal{S}$.
Note, that boundary velocities map to boundary derivatives w.r.t. $s$ by multiplying them with the total duration $T$, \ie $\bm{q}'_0 = T \dot{\bm{q}}_0$ and $\bm{q}'_T = T \dot{\bm{q}}_T$.
Furthermore, the optimisation problem in Eq.~\eqref{eq:obf} minimizes not only the objective $\bm{q}''(s)$, but also the integral over accelerations, since $\bm{q}''(s) = T^2 \ddot{\bm{q}}(s)$ and thus the objective $\int_0^{1} \ddot{\bm{q}}(s)^\trsp \ddot{\bm{q}}(s) ds$ directly maps to $\frac{1}{T^4} \int_0^{1} \bm{q}''(s)^\trsp \bm{q}''(s) ds$, corresponding to the control effort. It is minimal \textit{iff} the objective in Eq.~\eqref{eq:obf} is minimal. 
As a result, this trajectory representation provides a linear mapping from via points, boundary conditions and the movement duration to a time-continuous and smooth trajectory.

CC-VPSTO, analogously to VP-STO, exploits this explicit parameterisation with via-points and boundary conditions by optimizing \textit{only the via-points} while keeping the predefined boundary condition parameters fixed.

\section{Baseline for Offline Simulation Experiments: Derivation and Background}\label{sec:baseline}

We present an alternative approach to approximate the chance constraint in Eq.~\eqref{eq:cc_opt_basic} for the special case of obstacle collision avoidance, which we use as a baseline. 
For this, we leverage statistical learning theory \citep{shalev2014understanding,mohri2018foundations} to obtain an alternative confidence-based bound for $\kthres$, which we call $\krad$. This bound is more theoretically complete, as it does not require the independence of the Bernoulli variables, but we also note that it is more conservative, computationally expensive, and, less general since it is limited to a specific motion planning problem.

\subsection{Preliminaries on Statistical Learning Theory}
We remind the concept of \emph{Rademacher complexity} which is a measure used in statistical learning theory to quantify the complexity of a class of functions with respect to a given dataset.
The intuition is as follows: if the constraint function $g$ is ``simple'' (\eg a constant or linear function), then the complexity of the class $\calF$, defined later based on $g$, will be low.
As established by Proposition~\ref{prop:rademacher-general} in \citet{mohri2018foundations}, this implies that the ``generalization property'' of $\calF$ is good. In our case, this means that if a solution $\xb$ has a small rate of violating constraint $g$ with respect to $N$ i.i.d.~samples, then there is a good chance that the actual probability of constraint violation of $\xb$ is small too.

First, we introduce the notion of \emph{Rademacher complexity} and the associated \emph{generalization} result:

\begin{definition}
If $\calF$ is a (possibly infinite) set of functions from a set $\calD$ to $\Re$, \ie $\calF\subseteq\Re^\calD$, and $D=\{\deltab_1,\ldots,\deltab_N\}_{i=1}^N$ is a set of $N$ elements of $\calD$, then the \emph{Rademacher complexity of $\calF$ with respect to $D$} is defined by
\[
R_D(\calF) = E_{\sigmab_1,\ldots,\sigmab_N}\left[ \sup_{f\in\calF} \frac1N \sum_{i=1}^N \sigmab_i f(\deltab_i) \right],
\]
where $\{\sigmab_i\}_{i=1}^N$ are sampled independently uniformly at random in $\{-1,1\}$.
Furthermore, if $\Delta$ is a random variable with values in $\calD$, then the \emph{Rademacher complexity of $\calF$ with respect to $\Delta$ with $N$ samples} is defined by
\[
R_{\Delta,N}(\calF) = E_{\deltab_1\sim\Delta,\ldots,\deltab_N\sim\Delta}[R_D(\calF)]
\]
wherein $D$ stands for $\{\deltab_i\}_{i=1}^N$.
\end{definition}

A well-known result in statistical learning states that if $D=\{\deltab_i\}_{i=1}^N$ is a set of $N$ independent samples $\deltab_1\sim\Delta,\ldots,\deltab_N\sim\Delta$, then with confidence $1-\beta$ on the sampling of $D$, it holds that for every $f\in\calF$, $\frac1N\sum_{i=1}^N f(\deltab_i)$ is ``close'' to $E_{\deltab\sim\Delta}[f(\deltab)]$, where ``close'' is quantified with a quantity that depends on $\beta$, $N$ and $R_{\Delta,N}(\calF)$.
Formally, we have:

\begin{proposition}[{\citep[Theorem~3.3]{mohri2018foundations}}]\label{prop:rademacher-general}
It holds that
\begin{align}
&P_{\deltab_1\sim\Delta,\ldots,\deltab_N\sim\Delta}\bigg[\max_{f\in\calF}\Big\{E_{\deltab\sim\Delta}[f(\deltab)] - \frac1N\sum_{i=1}^Nf(\deltab_i)\Big\} \nonumber \\
&\hspace{0.5cm} {} \leq 2R_{\Delta,N}(\calF) + \sqrt{\frac{\log(\frac1\beta)}{2N}}\,\bigg] \geq 1-\beta.
\label{eq:pac-rademacher}
\end{align}
\end{proposition}

\subsection{Rademacher Complexity for Surrogate Constraint}
We apply the result in Proposition~\ref{prop:rademacher-general} to the surrogate optimisation problem in Eq.~\eqref{eq:cc_opt_surrogate}.
For that, we define $\calD$ as the domain of $\Delta$ and $\calF=\{\deltab\mapsto\boldsymbol{1}_{g(\xb,\deltab)\leq0}\mid\xb\in X\}\subseteq\Re^\calD$.
It holds that for each $\xb\in X$ and $D=\{\deltab_i\}_{i=1}^N\subseteq\calD$, $E_{\deltab\sim\Delta}[\boldsymbol{1}_{g(\xb,\deltab)\leq0}]=P[G_{\xb}=1]$ and $\sum_{i=1}^N \boldsymbol{1}_{g(\xb,\deltab_i)\leq0}=s_N(\xb;D)$.
Hence, we get the following:

\begin{corollary}\label{cor:pac-rademacher}
With $\calF$ defined as above, it holds that
\begin{align}
&P_{\deltab_1\sim\Delta,\ldots,\deltab_N\sim\Delta}\bigg[\max_{\xb\in X}\Big\{P[G_{\xb}=1] - \frac1N s_N(\xb;D)\Big\} \nonumber \\
&\hspace{0.5cm} {} \leq 2R_{\Delta,N}(\calF) + \sqrt{\frac{\log(\frac1\beta)}{2N}}\,\bigg] \geq 1-\beta,
\label{eq:pac-rademacher-us}
\end{align}
wherein $D$ stands for $\{\deltab_i\}_{i=1}^N$.
\end{corollary}

Corollary~\ref{cor:pac-rademacher} tells us that with confidence $1-\beta$ on the sampling of $D=\{\deltab_i\}_{i=1}^N$ with $N$ i.i.d.~samples from $\Delta$, \emph{any} solution $\xb$ that is feasible for the surrogate optimisation problem Eq.~\eqref{eq:cc_opt_surrogate} with
\begin{equation}\label{eq:pac-rademacher-us-k}
\kthres \leq \left(\eta - 2R_{\Delta,N}(\calF) - \sqrt{\frac{\log(\frac1\beta)}{2N}}\right)N
\end{equation}
is feasible for the original optimisation problem Eq.~\eqref{eq:cc_opt_basic}.

In view of Eq.~\eqref{eq:pac-rademacher-us-k}, computing a suitable $\kthres$ for the surrogate optimisation problem in Eq.~\eqref{eq:cc_opt_surrogate} can be approached by computing an upper bound on the Rademacher complexity of the associated set of functions $\calF$.
Despite the theoretical appeal of the this approach, computing an upper bound on the Rademacher complexity can be very challenging in general, and there is usually no closed-from expression for such bounds.
Nevertheless, we present here a tight bound for a special case of collision-avoidance problem.

\subsection{A Special Case of Collision Avoidance}\label{ssec:collision-avoidance}
We consider a robot motion planning problem, where a ball-shaped robot has to avoid $m$ ball-shaped obstacles with high probability across time instants $t_1,\ldots,t_H$.

For the sake of simplicity, we first focus on the case with one obstacle ($m=1$) and one time step ($H=1$), before generalising.
Thus, we consider the problem of finding the position $\xb\in X$ ($X\subseteq\Re^n$) of the center of the robot at time $t_1$ such that $P_{\deltab\sim\Delta}(\lVert\xb-\pb(\deltab)\rVert\geq r)\geq1-\eta$, where $r>0$ is the combined radius of the obstacle and the robot, and $\pb(\deltab)\in\Re^n$ is the position of the center of the obstacle at time $t_1$ under scenario $\deltab$.
In the formulation of Eq.~\eqref{eq:cc_opt_basic}, we have
\[
g(\xb,\deltab)=r-\lVert\xb-\pb(\deltab)\rVert,
\]
\ie $g(\xb,\deltab)\leq0\Leftrightarrow\lVert\xb-\pb(\deltab)\rVert\geq r$.

Let $\calF$ be defined as in the previous subsection, \ie $\calF=\{\boldsymbol{1}_{g(\xb,\deltab)\leq0}\mid\xb\in X\}$, with $X$ and $g$ as above.
In the subsequent section with the additional proofs (Appendix~\ref{sec:rademacher-proof}), we bound $R_{\Delta,N}(\calF)$ as follows:
\begin{equation}\label{eq:bound-rademacher-single}
R_{\Delta,N}(\calF)\leq\sqrt{\frac{d\log\left(\frac{eN}{d}\right)}{2N}},
\end{equation}
where $d=n+1$ and $e$ is Euler's number.
We obtain the following:

\begin{proposition}\label{prop:k-rademacher-single}
In the setting defined above with $m=H=1$, if we define $\krad(\beta,N,\eta)$ as
\[
\krad(\beta,N,\eta) = \eta N - \sqrt{2dN\log\left(\frac{eN}{d}\right)} - \sqrt{\frac{N\log(\frac1\beta)}{2}},
\]
then any feasible solution of the surrogate optimisation problem in Eq.~\eqref{eq:cc_opt_surrogate} with $\kthres=\krad(\beta,N,\eta)$ is feasible for the original optimisation problem in Eq.~\eqref{eq:cc_binom_version} with confidence $1-\beta$ on the sampling of $D$.
\end{proposition}

\begin{proof}
Consequence of Eqs.~\eqref{eq:pac-rademacher-us-k} and~\eqref{eq:bound-rademacher-single}.
\end{proof}

We now discuss the case of $m\in\Ne_{\geq1}$ obstacles and $H\in\Ne_{\geq1}$ time steps.
In this case,
\[
g(\xb,\deltab)=\max_{\substack{j=1,\ldots,m\\k=1,\ldots,H}}\,r-\lVert\qb(\xb,t_k)-\pb_j(\deltab,t_k)\rVert,
\]
\ie $g(\xb,\deltab)\leq0\Leftrightarrow\forall\,j\:\forall\,k\:\lVert\xb(t_k)-\pb_j(\deltab,t_k)\rVert\geq r$,
where $\xb(t_k)$ ($X\subseteq\Re^{nH}$) is the position of the center of the robot at time $t_k$ and $\pb_j(\deltab,t_k)\in\Re^n$ is the position of the center of the $j$\textsuperscript{th} obstacle at time $t_k$ under scenario $\deltab$. 
We show in Appendix~\ref{sec:rademacher-proof} that $R_{\Delta,N}(\calF)$ can be bounded as follows:
\begin{equation}\label{eq:bound-rademacher-multiple}
R_{\Delta,N}(\calF)\leq mH\sqrt{\frac{d\log\left(\frac{eN}{d}\right)}{2N}},
\end{equation}
where $d=n+1$ and $e$ is Euler's number.
Similarly to the above, we get the following:

\begin{proposition}\label{prop:k-rademacher-multiple}
In the setting defined above with $m\in\Ne_{\geq1}$ and $H\in\Ne_{\geq1}$, if we define $\krad(\beta,N,\eta)$ as
\begin{align*}
&\krad(\beta,N,\eta) = \\
&\hspace{10mm}\eta N - mH\sqrt{2dN\log\left(\frac{eN}{d}\right)} - \sqrt{\frac{N\log(\frac1\beta)}{2}},
\end{align*}
then any feasible solution of the surrogate optimisation problem in Eq.~\eqref{eq:cc_opt_surrogate} with $\kthres=\krad(\beta,N,\eta)$ is feasible for the original optimisation problem in Eq.~\eqref{eq:cc_binom_version} with confidence $1-\beta$ on the sampling of $D$.
\end{proposition}

\begin{proof}
Consequence of Eqs.~\eqref{eq:pac-rademacher-us-k} and~\eqref{eq:bound-rademacher-multiple}.
\end{proof}

\begin{remark}
Note that, unlike other approaches in the literature, Proposition~\ref{prop:k-rademacher-multiple} does not rely on \emph{Boole's inequality} to bound the joint probability of collision avoidance.
Indeed, the use of Boole's inequality would amount to set the collision avoidance probability for \emph{each time step and each obstacle} to $\eta'=\frac\eta{mH}$, so that the probability of collision \emph{with at least one obstacle at at least one time step} is bounded from above by $\eta=\sum_{j,k}\eta'$.
Furthermore, we would need to set the confidence for \emph{each time step and each obstacle} to $1-\beta'$ with $\beta'=\frac\beta{mH}$, in order to guarantee with confidence $1-\beta=1-\sum_{j,k}\beta'$ that the chance constraint holds for each of them simultaneously.
This would result in a value of $\kthres$ as follows:
\begin{align*}
&\kradbool(\beta,N,\eta) = \\
&\hspace{10mm}\frac{\eta N}{mH} - \sqrt{2dN\log\left(\frac{eN}{d}\right)} - \sqrt{\frac{N\log(\frac{mH}{\beta})}{2}}.
\end{align*}
This can be rewritten as
\begin{align*}
&\kradbool(\beta,N,\eta) = {} \\
&\hspace{10mm} \frac{\eta N - mH\sqrt{2dN\log\left(\frac{eN}{d}\right)} - mH\sqrt{\frac{N\log(\frac{mH}{\beta})}{2}}}{mH},
\end{align*}
The above shows that for any values of $\beta$, $N$ and $\eta$ for which $\krad(\beta,N,\eta)\geq0$, it holds that
\[
\kradbool(\beta,N,\eta)\leq\frac{1}{mH}\krad(\beta,N,\eta).
\]
Hence, using $\kthres=\krad(\beta,N,\eta)$ in Eq.~\eqref{eq:cc_opt_surrogate} is less conservative (by a ``factor'' $mH$) than using $\kthres=\kradbool(\beta,N,\eta)$.
\end{remark}

\subsection{Additional Proofs}\label{sec:rademacher-proof}
We start with Eq.~\eqref{eq:bound-rademacher-single}, reminded in the proposition below:

\begin{proposition}\label{prop:rademacher-single}
In the setting defined in Sec.~\ref{ssec:collision-avoidance} with $m=H=1$, it holds that
\begin{equation}\label{eq:rademacher-single}
R_{\Delta,N}(\calF) \leq \sqrt{\frac{d\log\left(\frac{eN}d\right)}{2N}},
\end{equation}
wherein $d=n+1$ and $e$ is Euler's number.
\end{proposition}

\begin{proof}
Consider the set of functions $\calH=\{2f-1\mid f\in\calF\}$, \ie
\begin{equation}\label{eq:class-H}
\calH=\{2\cdot\boldsymbol{1}_{\lVert\xb-\pb(\deltab)\rVert\leq r}-1\mid \xb\in X\},
\end{equation}
which is essentially the same as $\calF$ except that the functions take values in $\{-1,1\}$ instead of $\{0,1\}$,\footnote{This is done to stick to the classical theory of \emph{binary classification} learning for which many results on the Rademacher complexity have been derived \citep{shalev2014understanding,mohri2018foundations}.} and the quantity
\[
\Pi_\calH(N)=\max_{(\deltab_1,\ldots,\deltab_N)\in\calD^N}\#\{(h(\deltab_1),\ldots,h(\deltab_N))\mid h\in\calH\}.
\]
By \citep[Corollary~3.8]{mohri2018foundations}, it holds that
\[
R_{\Delta,N}(\calH)\leq\sqrt{\frac{2\log\Pi_\calH(N)}{N}}.
\]
We will bound $\Pi_\calH(N)$ by $\left(\frac{eN}{d}\right)^d$ by using Eq.~\eqref{eq:class-H}.
For that, we consider the set of functions $\calH'=\{h_{\xb,r}\mid(\xb,r)\in\Re^n\times\Re_{\geq0}\}\subseteq\Re^\calP$ where $\calP=\Re^n$ and $h_{\xb,r}(\pb)=2\cdot\boldsymbol{1}_{\lVert\xb-\pb\rVert\leq r}-1$, which contains all \emph{ball classifiers} in $\Re^n$ since $h_{\xb,r}(\pb)=1$ if $\pb$ is in the ball of centre $\xb$ and radius $r$, and $-1$ otherwise.
It follows that the VC dimension of $\calH'$, defined as the largest $N$ for which $\Pi_{\calH'}=2^N$, is $d=n+1$.\footnote{See for instance Sec.~15.5.2 in \url{https://ti.inf.ethz.ch/ew/lehre/CG12/lecture/Chapter\%2015.pdf} (last consulted: July 30, 2024).}
Hence, by \citep[Corollary~3.18]{mohri2018foundations}, it follows that $\Pi_{\calH'}(N)\leq\left(\frac{eN}{d}\right)^d$.
It is also straightforward to show that $\Pi_\calH(N)\leq\Pi_{\calH'}(N)$ since for every $\xb\in X$ and every $\deltab\in\calD$, $2\cdot\boldsymbol{1}_{\lVert\xb-\pb(\deltab)\rVert\leq r}-1=h_{\xb,r}(\pb(\deltab))$.
Hence,
\[
R_{\Delta,N}(\calH)\leq\sqrt{\frac{2d\log(eN/d)}{N}}.
\]
Finally, from \citep[Lemma~26.9]{shalev2014understanding}, we get that
\[
R_{\Delta,N}(\calF)\leq\frac12R_{\Delta,N}(\calH)\leq\sqrt{\frac{d\log(eN/d)}{2N}},
\]
concluding the proof.
\end{proof}

Next, we consider Eq.~\eqref{eq:bound-rademacher-multiple} and prove the following:

\begin{proposition}\label{prop:rademacher-multiple}
In the setting defined in Sec.\ref{ssec:collision-avoidance} with $m\in\Ne_{\geq1}$ and $H\in\Ne_{\geq1}$, it holds that
\begin{equation}\label{eq:rademacher-multiple}
R_{\Delta,N}(\calF) \leq mH\sqrt{\frac{d\log\left(\frac{eN}d\right)}{2N}},
\end{equation}
wherein $d=n+1$ and $e$ is Euler's number.
\end{proposition}

\begin{proof}
Note that by definition of $g$, it holds that
\begin{equation}\label{eq:max-class-F}
\boldsymbol{1}_{g(\xb,\deltab)>0} = \max_{j,k}\,\boldsymbol{1}_{\lVert\xb(t_k)-\pb_j(\deltab,t_k)\rVert\leq r}-1.
\end{equation}
For each $j=1,\ldots,m$ and $k=1,\ldots,H$, let
\[
\calF_{j,k}=\{\boldsymbol{1}_{\lVert\xb(t_k)-\pb_j(\deltab,t_k)\rVert\leq r}-1\mid\xb\in X\},
\]
and let $\calF'=\{\max_{j,k}\,f_{j,k} \mid \forall\,j\:\forall\,k\:f_{j,k}\in\calF_{j,k}\}$.
By Eq.~\eqref{eq:max-class-F}, it holds that $\calF\subseteq\calF'$.
By~\citep[Ex.~3.8]{mohri2018foundations}, it holds that $R_{\Delta,N}(\calF')=\sum_{j,k}R_{\Delta,N}(\calF_{j,k})$.
Furthermore, by Proposition~\ref{prop:rademacher-single}, we know that for each $j=1,\ldots,m$ and $k=1,\ldots,H$, $R_{\Delta,N}(\calF_{j,k})$ is bounded from above by the right-hand side of Eq.~\eqref{eq:rademacher-single}.
By summing over $j$ and $k$, we get that $R_{\Delta,N}(\calF')$ is bounded from above by the right-hand side term in Eq.~\eqref{eq:rademacher-multiple}.
Since $\calF\subseteq\calF'$, we have that $R_{\Delta,N}(\calF)\leq R_{\Delta,N}(\calF')$, concluding the proof.
\end{proof}

\begin{table}[t]
  \caption{Offline Planning Experiments for $\beta=0.05$}
  \centering
  \footnotesize
  \renewcommand{\arraystretch}{1.2}
  \begin{tabular}{|c|c|c|c|c|c|}
  \hline
  $\eta$ & $\eta_{\mathrm{rad}}$ & \begin{tabular}[c]{@{}c@{}}$\eta_{\mathrm{binom}}$\end{tabular} & $\hat{\eta}_{\mathrm{avg}}$ & \begin{tabular}[c]{@{}c@{}}$\hat {\eta}_{1-\beta}$\end{tabular} & $\bar{\beta}$ \\ \hline \hline
  \multirow{2}{*}{0.05} & n/a   & 0.01 & 0.0218 & 0.0499 & \multicolumn{1}{c|}{0.0497} \\ \cline{2-6} 
                        & n/a   & 0.038 & 0.0393 & 0.0503 & \multicolumn{1}{c|}{0.0539} \\ \hline \hline 
  \multirow{2}{*}{0.1}  & n/a   & 0.04 & 0.0525 & 0.0936 & \multicolumn{1}{c|}{0.0325} \\ \cline{2-6} 
                        & n/a   & 0.084 & 0.0854 & 0.1012 & \multicolumn{1}{c|}{0.0627} \\ \hline \hline 
  \multirow{2}{*}{0.15} & n/a   & 0.08 & 0.0934 & 0.1449 & \multicolumn{1}{c|}{0.0375} \\ \cline{2-6} 
                        & n/a   & 0.131 & 0.1322 & 0.1508 & \multicolumn{1}{c|}{0.0572} \\ \hline \hline
  \multirow{2}{*}{0.2}  & n/a   & 0.13 & 0.1438 & 0.2060 & \multicolumn{1}{c|}{0.0656} \\ \cline{2-6} 
                        & n/a   & 0.178 & 0.1794 & 0.2010 & \multicolumn{1}{c|}{0.0574} \\ \hline \hline
  \multirow{2}{*}{0.25} & n/a   & 0.17 & 0.1842 & 0.2515 & \multicolumn{1}{c|}{0.0535} \\ \cline{2-6} 
                        & 0.009 & 0.227 & 0.2288 & 0.2518 & \multicolumn{1}{c|}{0.0638} \\ \hline \hline
  \multirow{2}{*}{0.3}  & n/a   & 0.22 & 0.2350 & 0.3068 & \multicolumn{1}{c|}{0.0667} \\ \cline{2-6} 
                        & 0.059 & 0.275 & 0.2764 & 0.3005 & \multicolumn{1}{c|}{0.0533} \\ \hline \hline
  \multirow{2}{*}{0.35} & n/a   & 0.26 & 0.2755 & 0.3507 & \multicolumn{1}{c|}{0.0517} \\ \cline{2-6} 
                        & 0.109 & 0.324 & 0.3251 & 0.3510 & \multicolumn{1}{c|}{0.0578} \\ \hline \hline
  \multirow{2}{*}{0.4}  & n/a   & 0.31 & 0.3262 & 0.4053 & \multicolumn{1}{c|}{0.0627} \\ \cline{2-6} 
                        & 0.159 & 0.374 & 0.3760 & 0.4029 & \multicolumn{1}{c|}{0.0705} \\ \hline \hline
  \multirow{2}{*}{0.6}  & n/a   & 0.51 & 0.5282 & 0.6101 & \multicolumn{1}{c|}{0.0754} \\ \cline{2-6} 
                        & 0.359 & 0.573 & 0.5748 & 0.6025 & \multicolumn{1}{c|}{0.0683} \\ \hline \hline
  \multirow{2}{*}{0.8}  & 0.158 & 0.72 & 0.7359 & 0.8053 & \multicolumn{1}{c|}{0.0668} \\ \cline{2-6} 
                        & 0.559 & 0.778 & 0.7798 & 0.8024 & \multicolumn{1}{c|}{0.0714} \\ \hline
  \end{tabular}
  \label{tab:offline_analysis_results}
\end{table}

\section{Additional Details on Naive vs. Confidence-Bounded Surrogate Constraint}\label{sec:appendix_naive_vs_confidence}
\begin{figure}[t!]
  \centering 
  \includegraphics[width=\linewidth]{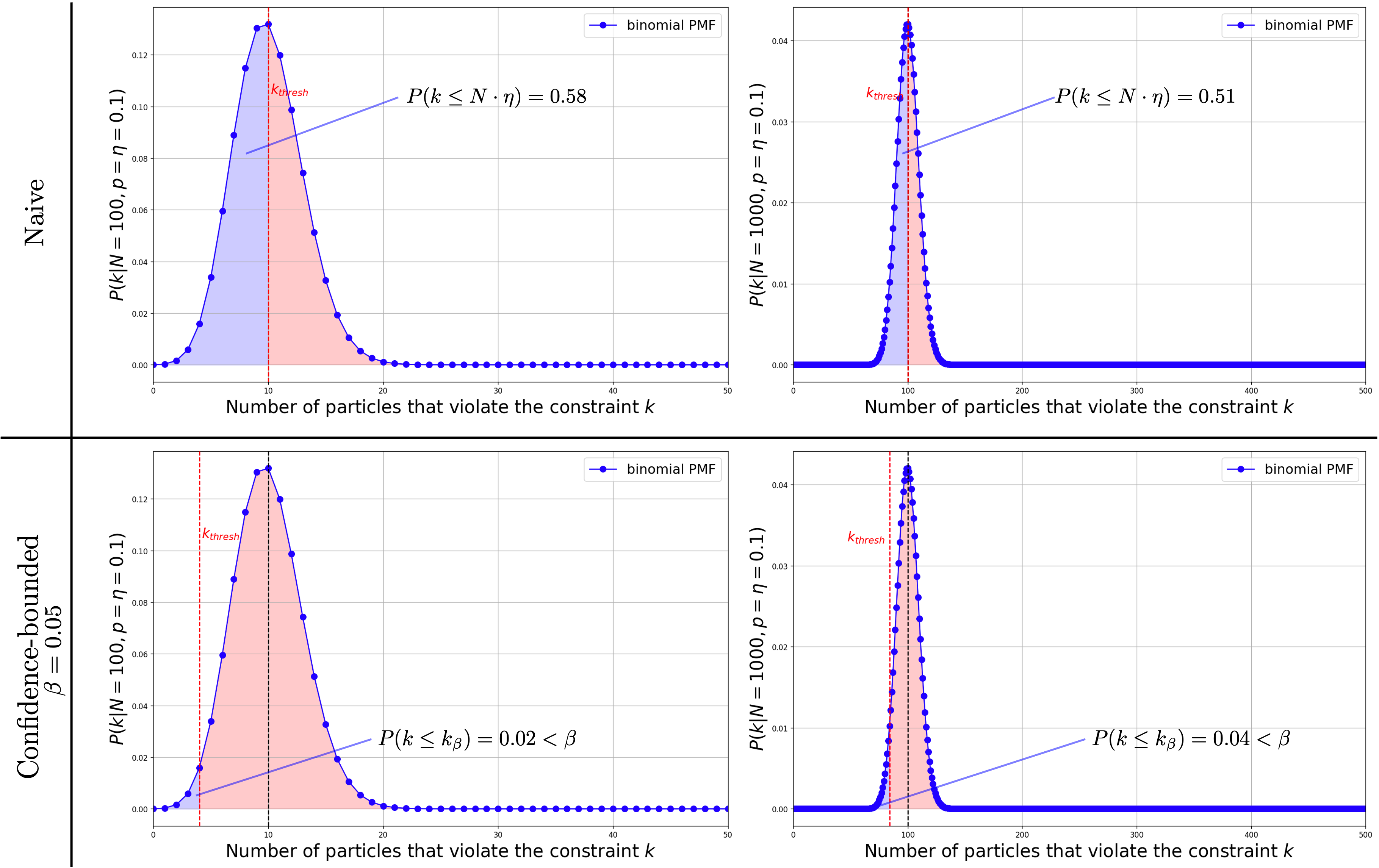}
  \caption{Analysis of the binomial distribution with $N=100$ (left column) and $N=1000$ (right column) Bernoulli experiments, which correspond to the number of samples used to approximate the chance constraint in the optimisation. 
  The top row shows the values $\kthres$ takes for the naïve formulation of setting $\kthres=\eta N$, where $\eta$ is the user-defined maximum probability of violation. The bottom row shows the values $\kthres$ takes for the confidence-bounded formulation of the chance constraint, \ie $\kthres=\kbinom(\beta,N,\eta)$ for $\beta=0.05$.}
  \label{fig:cc_binomial_analysis}
\end{figure}

We use Fig.~\ref{fig:cc_binomial_analysis} to illustrate the difference between the naïve formulation that only considers the maximum violation threshold $\eta$ by setting $\kthres=\eta N$ and the confidence-bounded formulation using $k_{\beta}$. For this purpose we analyse the binomial distribution with parameters $N$ and $p=\eta=0.1$ for different values of $N$, which correspond to the number of samples $\bm{\delta}_i \sim p_{\Delta}$ in the optimisation scheme. The plots in Fig.~\ref{fig:cc_binomial_analysis} shows the binomial distribution with $N=100$ on the left and $N=1000$ on the right. The blue shaded areas under the curve corresponds to the value of the CDF for $\kthres$, \ie $P(K\leq \kthres\mid N, \eta)$. The red-colored area under the curve corresponds to the the probability that $k>\kthres$. The top row shows the naïve formulation, \ie setting $\kthres=\eta N$. The bottom row shows the confidence-bounded formulation, \ie setting $\kthres=\kbinom(\beta,N,p)$ for $\beta=0.05$.

\begin{table*}[t]
  \caption{MPC Environment Specifications}
  \centering
  \scriptsize
    \setlength{\tabcolsep}{1pt} 
      \renewcommand{\arraystretch}{1.1}
      \begin{tabular}{|c|c|c|c|} \hline 
      Env.& 0& 1&2\\ \hline \hline 
      $N_{obs}$ &  5&  4&  5\\ \hline 
      Robot radius & 0.25 & 0.5 & 0.5 \\ \hline
      $\bm{x}_0$&   $\left[ \begin{array}{c}
        \begin{bmatrix} 2.0 \\ 4.0 \end{bmatrix}, 
        \begin{bmatrix} 3.5 \\ 8.0 \end{bmatrix}, 
        \begin{bmatrix} 7.5 \\ 2.5 \end{bmatrix}, 
        \begin{bmatrix} 9.0 \\ 1.5 \end{bmatrix}, 
        \begin{bmatrix} 4.5 \\ 8.0 \end{bmatrix}
        \end{array} \right]$& $\left[ \begin{array}{c}
        \begin{bmatrix} 7.9 \\ 5.7 \end{bmatrix}, 
        \begin{bmatrix} 1.3 \\ 3.5 \end{bmatrix}, 
        \begin{bmatrix} 4.9 \\ 9.4 \end{bmatrix}, 
        \begin{bmatrix} 5.2 \\ 3.0 \end{bmatrix}
        \end{array} \right]$ & $\left[ \begin{array}{c}
        \begin{bmatrix} 2.1 \\ 3.1 \end{bmatrix}, 
        \begin{bmatrix} 6.8 \\ 5.0 \end{bmatrix}, 
        \begin{bmatrix} 7.3 \\ 6.7 \end{bmatrix}, 
        \begin{bmatrix} 4.2 \\ 4.2 \end{bmatrix},
        \begin{bmatrix} 8.5 \\ 2.8 \end{bmatrix}
        \end{array} \right]$  \\ \hline 
      $\bm{\dot{x}}_0$ & $\left[ \begin{array}{c}
        \begin{bmatrix} 0.7 \\ 0.0 \end{bmatrix}, 
        \begin{bmatrix} 0.25 \\ -0.5 \end{bmatrix}, 
        \begin{bmatrix} -0.5 \\ 0.5 \end{bmatrix}, 
        \begin{bmatrix} -0.1 \\ 0.1 \end{bmatrix}, 
        \begin{bmatrix} 0.0 \\ -1.0 \end{bmatrix}
        \end{array} \right]$ & $\left[ \begin{array}{c}
        \begin{bmatrix} 0.6 \\ 0.1 \end{bmatrix}, 
        \begin{bmatrix} 0.0 \\ 0.2 \end{bmatrix}, 
        \begin{bmatrix} -0.4 \\ 0.1 \end{bmatrix}, 
        \begin{bmatrix} -0.2 \\ 0.0 \end{bmatrix}
        \end{array} \right]$  &  $\left[ \begin{array}{c}
        \begin{bmatrix} 0.5 \\ -0.2 \end{bmatrix}, 
        \begin{bmatrix} 0.5 \\ 0.0 \end{bmatrix}, 
        \begin{bmatrix} 0.0 \\ -0.2 \end{bmatrix}, 
        \begin{bmatrix} 0.4 \\ 0.6 \end{bmatrix}, 
        \begin{bmatrix} 0.2 \\ -0.3 \end{bmatrix}
        \end{array} \right]$ \\ \hline
      Radii & $[0.5, 0.4, 0.3, 0.35, 0.55]$ & $[- 0.32, 0.51, 0.49, 0.34] $& $[0.54, 0.45, 0.55, 0.35, 0.34]$\\ \hline  
      $var(\ddot{\bm{x}})$ & $[0.5, 0.75, 0.65, 0.8, 0.6]$ & $[0.54, 0.64, 0.51, 0.8]$ & $[0.64, 0.66, 0.62, 0.57, 0.75]$\\ \hline 
  \end{tabular}
  \label{tab:env_specs}
\end{table*}

\section{Experiment Details}\label{sec:appendix-experiments}
\subsection{MPC Experiments: Environment Details}\label{sec:mpc_env_details}
In this section, we provide additional details about the environments used for the MPC experiments in Sec.~\ref{sec:simulation_experiments}. We used three different environment configurations for which we generated the parameters randomly. Tab.~\ref{tab:env_specs} shows the specifications of the environments, \ie the number of obstacles $N_{obs}$, the initial obstacle positions $\bm{x}_0$, the initial obstacle velocities $\bm{\dot{x}}_0$, the obstacle radii and the variance of the obstacle accelerations $\ddot{\bm{x}}$, when sampling from a zero-mean Gaussian distribution in the random-walk model. The environment size was chosen to be consistent across all environments on a 10 by 10 grid. 

\subsection{Detailed Results on Offline-CC-VPSTO}\label{sec:appendix_offline_cc_vpsto}
The numerical results for the offline planning experiments are shown in Tab.~\ref{tab:offline_analysis_results}.

\subsection{Robot Experiment: Implementation of Stochastic Model}\label{sec:robot_exp_details}
In this section, we provide additional details about the implementation of the stochastic model for the robot experiment in Sec.~\ref{sec:robot_experiment} describing the motion of the box obstacle on the conveyor belt. As our approach is Monte Carlo-based, in every MPC step we simulate the motion of the box obstacle for $N_{\text{sim}}$ samples. The samples are initialised with the same position and velocity as the box obstacle at the beginning of the MPC step. The samples are then propagated through the conveyor belt dynamics for the duration of the MPC step. The conveyor belt dynamics are modelled as a probabilistic system. 
A sample at time step $k$ is modeled by state vector $\mathbf{s} = [x_k, \dot{x}_k, p_k]$ where $x_k$ is the position, $\dot{x}_k$ is the velocity, and $p_k$ is the probability of changing direction at time step $k$

The direction-change mechanism works as follows. At each time 
step, the probability $p_k$ is first decayed by the update rule 
$p_{k+1} = p_k \cdot (1 - \alpha)$, and then a random number 
$r \sim \mathcal{U}(0,1)$ is sampled. A direction change is triggered if $r < p_{k+1}$ or if the projected position $x_k + \dot{x}_k \Delta t$ lies outside the conveyor belt boundaries. Upon a direction change, $p_k$ is reset to $\alpha$. Since $p_k$ decays geometrically as $\alpha(1-\alpha)^k$ (where $k$ counts steps since the last reset), this model implements a  \emph{decreasing hazard rate}: direction changes are most probable shortly after a previous change and become progressively less likely as the box continues moving in the same direction. This captures the intuition that once the box has been travelling consistently in one direction, a sudden reversal becomes less likely. The parameter $\alpha$ controls the overall frequency of 
direction changes: higher values of $\alpha$ lead to more frequent reversals.

The dynamics of this system for each time step $k$ can be described as follows:

\begin{enumerate}
    \item \textit{Update the Probability of Direction Change:}
    \begin{equation}
        p_{k+1} = p_{k} \cdot (1 - \alpha)
    \end{equation}
    where $\alpha$ controls the decay rate of the direction-change probability.

    \item \textit{Determine the Direction Change:}
    \begin{itemize}
        \item Sample a random number $r$ from a uniform distribution between 0 and 1.
        \item If $r < p_{k+1}$ or if the projected position $x_k + \dot{x}_k \Delta t$ is outside the boundaries of the conveyor belt, a direction change occurs.
    \end{itemize}

    \item \textit{Update State based on Direction Change:}
    \begin{equation}
        \dot{x}_{k+1} =
        \begin{cases}
            -\dot{x}_k & \text{if direction change occurs} \\
            \dot{x}_k & \text{otherwise}
        \end{cases}
    \end{equation}
    \begin{equation}
        p_{k+1} =
        \begin{cases}
            \alpha & \text{if direction change occurs} \\
            p_{k+1} & \text{otherwise}
        \end{cases}
    \end{equation}

    \item \textit{Update Position:}
    \begin{equation}
        x_{k+1} = x_k + \dot{x}_{k+1} \Delta t
    \end{equation}
\end{enumerate}

Therefore, the updated state vector after each time step is:
\begin{equation}
    \mathbf{s}_{k+1} = [x_{k+1}, \, \dot{x}_{k+1}, \, p_{k+1}]
\end{equation}

In summary, the above models the probabilistic dynamics of one box sample on the conveyor belt, where the direction of motion can change randomly influenced by the parameter $\alpha$ and the physical constraints of the system.

\end{document}